\documentclass[10pt,twocolumn,letterpaper]{article}
\usepackage{titling}
\usepackage{iccv}
\usepackage{times}
\usepackage{epsfig}
\usepackage{graphicx}
\usepackage{array,multirow}
\usepackage{float}
\usepackage{amsmath}
\usepackage{amssymb}
\usepackage{graphics}
\usepackage{subfig}
\usepackage{color}
\usepackage{caption}
\usepackage[english]{babel}
\usepackage{booktabs} 
\usepackage[ruled,vlined,commentsnumbered,linesnumbered]{algorithm2e} 
\usepackage[dvipsnames]{xcolor}

\newcommand{\PAR}[1]{\vskip4pt \noindent{\bf #1~}}

\usepackage{fancyhdr}
\usepackage{setspace}

\usepackage[pagebackref=true,breaklinks=true,letterpaper=true,colorlinks,bookmarks=false]{hyperref}
\newif\ifarxiv
\arxivtrue

\ifarxiv
    \iccvfinalcopy
\fi

\iccvfinalcopy 

\def\iccvPaperID{2441} 

\ificcvfinal\fi

\def\paperTitle{Tracking without bells and whistles}

\begin{document}

\title{\paperTitle}


\author{\qquad Philipp Bergmann%
\thanks{Contributed equally. Correspondence to: tim.meinhardt@tum.de}\qquad
\and \qquad Tim Meinhardt
\footnotemark[1]\qquad
\and \qquad Laura Leal-Taixe\qquad
\vspace{0.2cm}
\and
\qquad \normalfont{\textit{Technical University of Munich}}\qquad
}


\maketitle
\ifarxiv
    \thispagestyle{fancy}
\fi

\global\csname @topnum\endcsname 0
\global\csname @botnum\endcsname 0

\begin{abstract}
The problem of tracking multiple objects in a video sequence poses several challenging tasks.
For tracking-by-detection, these include object re-identification, motion prediction and dealing with occlusions.
We present a tracker (without bells and whistles) that accomplishes tracking without specifically targeting any of these tasks, in particular, we perform no training or optimization on tracking data.
To this end, we exploit the bounding box regression of an object detector to predict the position of an object in the next frame, thereby converting a detector into a {\it Tracktor}.
We demonstrate the potential of Tracktor and provide a new state-of-the-art on three multi-object tracking benchmarks by extending it with a straightforward re-identification and camera motion compensation.

We then perform an analysis on the performance and failure cases of several state-of-the-art tracking methods in comparison to our Tracktor.
Surprisingly, none of the dedicated tracking methods are considerably better in dealing with complex tracking scenarios, namely, small and occluded objects or missing detections.
However, our approach tackles most of the easy tracking scenarios.
Therefore, we motivate our approach as a new tracking paradigm and point out promising future research directions.
Overall, {\it Tracktor} yields superior tracking performance than any current tracking method and our analysis exposes remaining and unsolved tracking challenges to inspire future research directions.

%
%
\end{abstract}


\vspace{-0.5cm}
\section{Introduction}


\begin{figure*}[t!]
    \centering
    \includegraphics[width=\textwidth]{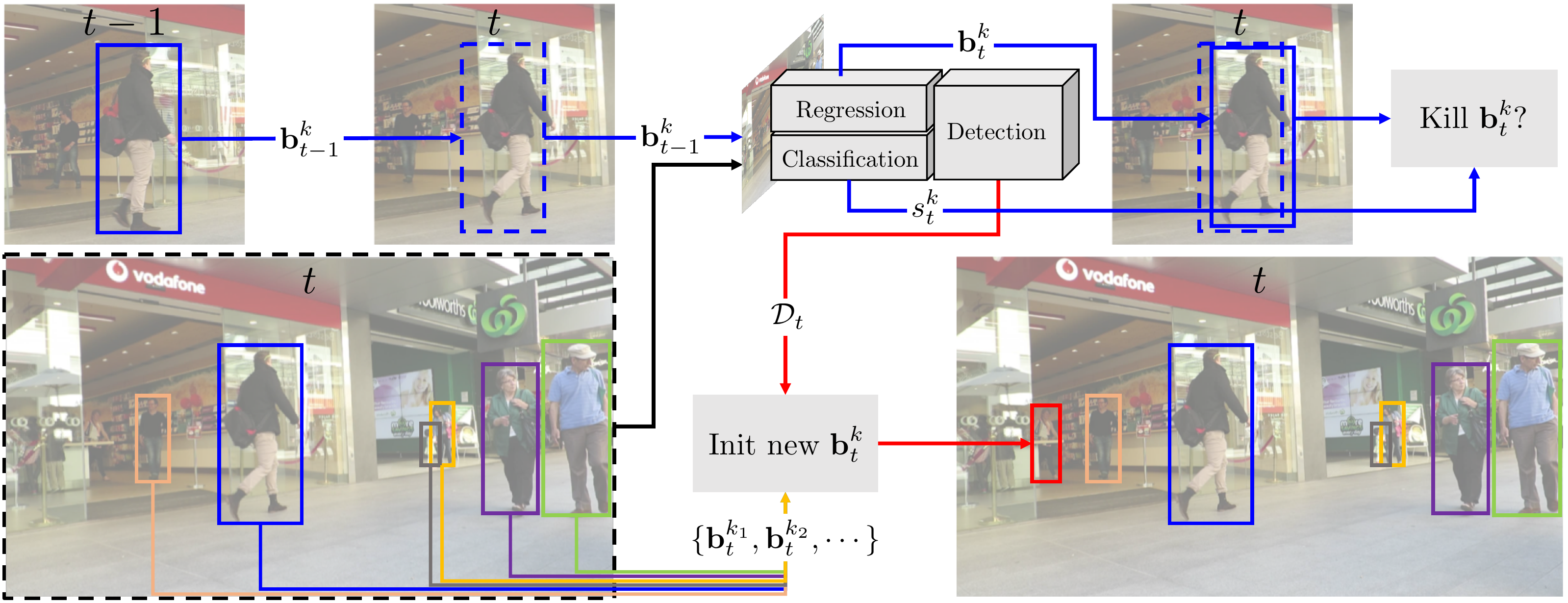}
    \caption{%
    The presented Tracktor accomplishes multi-object tracking only with an object detector and consists of two primary processing steps, indicated in blue and red, for a given frame $t$.
    First, the regression of the object detector aligns already existing track bounding boxes ${\bf b}^k_{t-1}$ of frame $t-1$ to the object's new position at frame $t$.
    The corresponding object classification scores $s^k_t$ of the new bounding box positions are then used to kill potentially occluded tracks.
    Second, the object detector (or a given set of public detections) provides a set of detections $\mathcal{D}_t$ of frame $t$.
    Finally, a new track is initialized if a detection has no substantial Intersection over Union with any bounding box of the set of active tracks $B_t=\{{{\bf b}^{k_1}_t},{{\bf b}^{k_2}_t}, \cdots\}$.}
    \label{fig:method_vis}
\end{figure*}

Scene understanding from video remains one of the big challenges of computer vision.
Humans are often the center of attention in a scene, which leads to the fundamental problem of detecting and tracking them in a video.
\emph{Tracking-by-detection} has emerged as the preferred paradigm to solve the problem of tracking multiple objects as it simplifies the task by breaking it into two steps: (i) detecting object locations independently in each frame, (ii) form tracks by linking corresponding detections across time.
The linking step, or \emph{data association}, is a challenging task on its own, due to missing and spurious detections, occlusions, and target interactions in crowded environments.
To address these issues, research in this area has produced increasingly complex models achieving only marginally better results, e.g., multiple object tracking accuracy has only improved 2.4\% in the last two years on the MOT16~\cite{milanarxiv2016} benchmark.


In this paper, we push tracking-by-detection to the limit by using only an object detection method to perform tracking.
We show that one can achieve state-of-the-art tracking results by training a neural network only on the task of {\it detection}. 
As indicated by the blue arrows in Figure~\ref{fig:method_vis}, the regressor of an object detector such as Faster-RCNN~\cite{rennips2015} is sufficient to construct object trajectories in a multitude of challenging tracking scenarios.
%
%
This raises an interesting question that we discuss in this paper: If a {\it detector} can solve most of the tracking problems, what are the real situations where a dedicated {\it tracking} algorithm is necessary?
%
%
We hope our work and the presented {\it Tracktor} allows researchers to focus on the still unsolved critical challenges of multi-object tracking.


This paper presents four main  contributions:
\begin{itemize}
\item{We introduce the Tracktor which tackles multi-object tracking by exploiting the regression head of a detector to perform temporal realignment of object bounding boxes.}
\item{We present two simple extensions to Tracktor, a re-identification Siamese network and a motion model. The resulting tracker yields state-of-the-art performance in three challenging multi-object tracking benchmarks.}
\item{We conduct a detailed analysis on failure cases and challenging tracking scenarios, and show none of the dedicated tracking methods perform substantially better than our regression approach.}
\item{We propose our method as a new tracking paradigm which exploits the detector and allows researchers to focus on the remaining complex tracking challenges. This includes an extensive study on promising future research directions.}
%
%
\end{itemize}

\subsection{Related work}

Several computer vision tasks such as surveillance, activity recognition or autonomous driving rely on object trajectories as input.
Despite the vast literature on multi-object tracking~\cite{Luo:2014:arXiv, LealTaix2017TrackingTT}, it still remains a challenging problem, especially in crowded environments where occlusions and false detections are common.
%
Most state-of-the-art works follow the tracking-by-detection paradigm which heavily relies on the performance of the underlying detection method.

%
Recently, neural network based detectors have clearly outperformed all other methods for detection~\cite{krizhevskynips2012, rennips2015,yolov2}.
%
%
The family of detectors that evolved to Faster-RCNN~\cite{rennips2015}, and further detectors such as SDP~\cite{sdpYangcvpr2016}, rely on object proposals which are passed to an object classification and a bounding box regression head of a neural network. The latter refines bounding boxes to fit tightly around the object.
In this paper, we show that one can rethink the use of this regressor for tracking purposes.

\noindent{\bf Tracking as a graph problem.}
The data association problem deals with keeping the identity of the tracked objects given the available detections.
This can be done on a frame by frame basis for online applications \cite{breitensteiniccv2009, esscvpr2008, pellegriniiccv2009} or track-by-track \cite{berclazcvpr2006}.
Since video analysis can be done offline, batch methods are preferred since they are more robust to occlusions.
A common formalism is to represent the problem as a graph, where each detection is a node, and edges indicate a possible link.
The data association can then be formulated as maximum flow~\cite{berclaztpami2011} or, equivalently, minimum cost problem with either fixed costs based on distance~\cite{jiangcvpr2007, pirsiavashcvpr2011,zhangcvpr2008}, including motion models~\cite{lealiccv2011}, or learned costs~\cite{lealcvpr2014}.
%
%
Alternative formulations typically lead to more involved optimization problems, including minimum cliques~\cite{zamireccv2012}, general-purpose solvers like MCMC~\cite{yucvpr2007} or multi-cuts~\cite{TangAAS17}.
%
%
A recent trend is to design ever more complex models which include other vision input such as reconstruction for multi-camera sequences~\cite{lealcvpr2012,wucvpr2011}, activity recognition~\cite{choieccv2012}, segmentation~\cite{milancvpr2015}, keypoint trajectories~\cite{choiiccv2015} or joint detection~\cite{TangAAS17}.
In general, the significantly higher computational costs do not translate to significantly higher accuracy.
In fact, in this work, we show that we can outperform all graph-based trackers significantly while keeping the tracker online.
Even within a graphical model optimization, one needs to define a measure to identify whether two bounding boxes belong to the same person or not.
This can be done by analyzing either the appearance of the pedestrian, or its motion.

\noindent{\bf Appearance models and re-identification.}
%
Discriminating and re-identifying (reID) objects by appearance is in particular a problem in crowded scenes with many object-object occlusions.
%
In the exhaustive literature that uses appearance models or reID methods to improve multi-object tracking, color-based models are very common~\cite{kimiccv2015}.
However, these are not always reliable for pedestrian tracking, since people can wear very similar clothes, and color statistics are often contaminated by background pixels and illumination changes.
The authors of~\cite{kuocvpr2011} borrow ideas from person re-identification and adapt them to ``re-identify'' targets during tracking.
In~\cite{yangcvpr2012}, a CRF model is learned to better distinguish pedestrians with similar appearance. 
Both appearance and short-term motion in the form of optical flow can be used as input to a Siamese neural network to decide whether two boxes belong to the same track or not~\cite{lealcvprw2016}. 
Recently,~\cite{ristanicvpr2018} showed the importance of learned reID features for multi-object tracking. We confirm this view in our experiments.


\noindent{\bf Motion models and trajectory prediction.} 
Several works resort to motion to discriminate between pedestrians, especially in highly crowded scenes.
%
%
The most common assumption is the one of constant velocity (CVA)~\cite{choieccv2010,andriyenkocvpr2011}, but pedestrian motion gets more complex in crowded scenarios for which researchers have turned to the more expressive Social Force Model~\cite{scovannericcv2009, pellegriniiccv2009, yamaguchicvpr2011, lealiccv2011}.
Such a model can also be learned from data~\cite{lealcvpr2014}.
Deep Learning has been extensively used to learn social etiquette in crowded scenarios for trajectory prediction~\cite{lealiccv2011, alahicvpr2016,robicqueteccv2016}.
\cite{zhueccv2018} use single object tracking trained networks to create tracklets for further post-processing into trajectories.
Recently,~\cite{Chen_2018_ECCV, reneccv2018} proposed to use reinforcement learning to predict the position of an object in the next frame.
While~\cite{Chen_2018_ECCV} focuses on single object tracking, the authors of~\cite{reneccv2018} train a multi-object pedestrian tracker composed of a bounding box predictor and a decision network for collaborative decision making between tracked objects.
%

\noindent{\bf Video object detection.}
Multi-object tracking without frame-to-frame identity prediction is a subproblem usually referred to as video object detection.
In order to improve detections, many methods exploit spatio-temporal consistencies of object positions. 
Both~\cite{Kang2016ObjectDF} and~\cite{Kang2017ObjectDI} generate multi-frame bounding box tuplet proposals and extract detection scores and features with a CNN and LSTM, respectively.
Recently, the authors of~\cite{Ogden2018TCNNTW} improve object detections by applying optical flow to propagate scores between frames.
Eventually,~\cite{Feichtenhofer2017DetectTT} proposes to solve the tracking and detection problem jointly.
They propose a network which processes two consecutive frames and exploits tracking ground truth data to improve detection regression, thereby, generating two-frame tracklets.
With a subsequent offline method, these tracklets are combined to multi-frame tracks.
However, we show that our regression tracker is not only online, but superior in dealing with object occlusions.
In particular, we do not only temporally align detections, but preserve their identity.


\section{A detector is all you need}
%

%
We propose to convert a {\it detector} into a {\it Tracktor} performing multiple object tracking.
Several CNN-based detection algorithms~\cite{rennips2015,sdpYangcvpr2016} contain some form of bounding box refinement through regression. 
We propose an exploitation of such a regressor for the task of tracking.
This has two key advantages: (i) we do not require any tracking specific training, and (ii) we do not perform any complex optimization at test time, hence our tracker is online.
Furthermore, we show that our method achieves state-of-the-art performance on several challenging tracking scenarios.
%
%

\subsection{Object detector}
The core element of our tracking pipeline is a regression-based detector.
In our case, we train a Faster R-CNN~\cite{rennips2015} with ResNet-101~\cite{HeZRS15} and Feature Pyramid Networks (FPN)~\cite{FPN_2017_CVPR} on the MOT17Det~\cite{milanarxiv2016} pedestrian detection dataset.

%
%
%
To perform object detection, Faster R-CNN applies a Region Proposal Network to generate a multitude of bounding box proposals for each potential object.
Feature maps for each proposal are extracted via Region of Interest (RoI) pooling~\cite{Girshick2015FastR}, and passed to the classification and regression heads.
The classification head assigns an object score to the proposal, in our case, it evaluates the likelihood of the proposal showing a pedestrian.
The regression head refines the bounding box location tightly around an object.
%
%
%
The detector yields the final set of object detections by applying non-maximum-suppression (NMS) to the refined bounding box proposals.
Our presented method exploits the aforementioned ability to regress and classify bounding boxes to perform multi-object tracking.

\subsection{Tracktor}
\label{sec:tracktor}

The challenge of multi-object tracking is to extract the spatial and temporal positions, i.e., trajectories, of $k$ objects given a frame by frame video sequence.
Such a trajectory is defined as a list of ordered object bounding boxes $T_k=\{{{\bf b}^k_{t_1}},{{\bf b}^k_{t_2}}, \cdots \}$, where a bounding box is defined by its coordinates ${\bf b}^k_t=(x,y,w,h)$, and $t$ represents a frame of the video.
We denote the set object bounding boxes in frame $t$ with $B_t=\{{{\bf b}^{k_1}_t},{{\bf b}^{k_2}_t}, \cdots\}$.
Note, that each $T_k$ or $B_t$ can contain less elements than the total number of frames or trajectories in a sequence, respectively.
%
%
At $t=0$, our tracker initializes tracks from the first set of detections $\mathcal{D}_0=\{{\bf d}^1_0, {\bf d}^2_0, \cdots\}=B_0$.
In Figure~\ref{fig:method_vis}, we illustrate the two subsequent processing steps (the nuts and bolts of our method) for a given frame $t$ for all $t > 0$, namely, the bounding box regression and track initialization.

\noindent{\bf Bounding box regression.}
The first step, denoted with blue arrows, exploits the bounding box regression to extend active trajectories to the current frame $t$.
This is achieved by regressing the bounding box ${\bf b}^k_{t-1}$ of frame $t-1$ to the object's new position ${\bf b}^k_t$ at frame $t$. 
In the case of Faster R-CNN, this corresponds to applying RoI pooling on the features of the current frame but with the previous bounding box coordinates.
Our assumption is that the target has moved only slightly between frames, which is usually ensured from high frame rates (see Section B.5 of the supplementary for a frame rate robustness evaluation of Tracktor).
The identity is automatically transferred from the previous to the regressed bounding box, effectively creating a trajectory.
This is repeated for all subsequent frames.

After the bounding box regression, our tracker considers two cases for killing (deactivating) a trajectory: (i) an object leaving the frame or occluded by a non-object is killed if its new classification score $s^k_t$ is below $\sigma_{active}$ and (ii) occlusions between objects are handled by applying non-maximum suppression (NMS) to all remaining $B_t$ and their corresponding scores with an Intersection over Union (IoU) threshold $\lambda_{active}$.



\noindent{\bf Bounding box initialization.}
In order to account for new targets, the object detector also provides the detections $\mathcal{D}_t$ for the entire frame $t$.
This second step, indicated in Figure~\ref{fig:method_vis} with red arrows, is analogous to the first initialization at $t=0$.
But a detection from $\mathcal{D}_t$ starts a trajectory only if the IoU with any of the already active trajectories ${\bf b}^k_t$ is smaller than $\lambda_{new}$.
That is, we consider a detection for a new trajectory only if it is covering a potentially new object that is not explained by any trajectory.
%
It should be noted again that our Tracktor does not require any tracking specific training or optimization and solely relies on an object detection method.
This allows us to directly benefit from improved object detection methods and, most importantly, enables a comparatively cheap transfer to different tracking datasets or scenarios in which no ground truth tracking but only detection data is available.

\subsection{Tracking extensions}
%
%
In this section, we present two straightforward extensions to our vanilla Tracktor: a motion model and a re-identification algorithm.
Both are aimed at improving identity preservation across frames and are common examples of techniques used to enhance, e.g., graph-based tracking methods~\cite{lealiccv2011, yangcvpr2012, lealcvprw2016}.
%
%
%

\noindent{\bf Motion model.}
%
Our previous assumption that the position of an object changes only slightly from frame to frame does not hold in two scenarios: large camera motion and low video frame rates.
In extreme cases, the bounding boxes from frame ${t-1}$ might not contain the tracked object in frame $t$ at all.
%
Therefore, we apply two types of motion models that will improve the bounding box position in future frames.
For sequences with a moving camera, we apply a straightforward camera motion compensation (CMC) by aligning frames via image registration using the Enhanced Correlation Coefficient (ECC) maximization as introduced in~\cite{EvangelidisP08}.
%
%
%
For sequences with comparatively low frame rates, we apply a constant velocity assumption (CVA) for all objects as in~\cite{choieccv2010,andriyenkocvpr2011}.


\noindent{\bf Re-identification.}
%
%
%
%
In order to keep our tracker online, we suggest a short-term re-identification (reID) based on appearance vectors generated by a Siamese neural network~\cite{bromleySiamese1993nips, HermansBL17,ristanicvpr2018}. 
To that end, we store killed (deactivated) tracks in their non-regressed version ${\bf b}^k_{t-1}$ for a fixed number of $F_{reID}$ frames.
%
We then compare the distance in the embedding space of the deactivated with the newly detected tracks and re-identify via a threshold.
The embedding space distance is computed by a Siamese CNN and appearance feature vectors for each of the bounding boxes.
It should be noted that the reID network is indeed trained on tracking ground truth data.
%
%
To minimize the risk of false reIDs, we only consider pairs of deactivated and new bounding boxes with a sufficiently large IoU.
The motion model is continuously applied to the deactivated tracks.


\section{Experiments}
We demonstrate the tracking performance of our proposed Tracktor tracker as well as its extension {\it Tracktor++} on several datasets focusing on pedestrian tracking.~\footnote{Tracktor code: ~\url{https://git.io/fjQr8}.} 
In addition, we perform an ablation study of the aforementioned extensions and further show that our tracker outperforms state-of-the-art methods in tracking accuracy and excels at identity preservation.
%


\noindent{\bf MOTChallenge.}
The multi-object tracking benchmark MOTChallenge~\footnote{The MOTChallenge web page: ~\url{https://motchallenge.net}.} consists of several challenging pedestrian tracking sequences, with frequent occlusions and crowded scenes.
Sequences vary in their angle of view, size of objects, camera motion and frame rate.
The challenge contains three separate tracking benchmarks, namely {\it 2D MOT 2015}~\cite{lealarxiv2015}, {\it MOT16} and {\it MOT17}~\cite{milanarxiv2016}.
The MOT17 test set includes a total of 7 sequences each of which is provided with three sets of public detections.
The detections originate from different object detectors each with increasing performance, namely DPM~\cite{dpmpami2009}, Faster R-CNN~\cite{rennips2015} and SDP~\cite{sdpYangcvpr2016}.
Our object detector is trained on the MOT17Det~\cite{milanarxiv2016} detection benchmark which contains the same images as MOT17.
The MOT16 benchmark also contains the same sequences as MOT17 but only provides DPM public detections.
The 2D MOT 2015 benchmark provides ACF~\cite{acfpami2014} detections for 11 sequences.
The complexity of the tracking problem requires several metrics to measure different aspects of a tracker's performance.
The Multiple Object Tracking Accuracy (MOTA)~\cite{clear} and ID F1 Score (IDF1)~\cite{ristanieccvw2016} quantify two of the main aspects, namely, object coverage and identity.
%

%
%
%
%
%

\noindent{\bf Public detections.}
%
%
For a fair comparison with other tracking methods, we perform all experiments with the public detections provided by MOTChallenge.
That is, all methods compared in this paper, including our approach and its extension, process the same precomputed frame by frame detections.
%
For our method, a new trajectory is {\it only} initialized from a public detection bounding box, i.e., we {\it never} use our object detector to detect a new bounding box.
We only apply the bounding box regressor and classifier to obtain new ${\bf b}^k_t$ and $s^k_t$, respectively.
The MOTChallenge public benchmark includes multiple methods~\cite{jCCpami2018,ChenAZS18,yoonavss2018} which classify the given detections with trained neural networks, hence, we consider our processing of the given detections also as {\it public}.
%
%
%
%
%
%
%

\subsection{Ablation study}
\label{sec:ablation}


\begin{table}
\center 
\tabcolsep=0.11cm

    \resizebox{\columnwidth}{!}{
    \begin{tabular}{l c c c c c c c}
        \toprule
        Method & MOTA $\uparrow$ & IDF1 $\uparrow$ & MT $\uparrow$ & ML $\downarrow$ & FP $\downarrow$ & FN $\downarrow$ & ID Sw. $\downarrow$ \\ [0.5ex] 
        \midrule
        D\&T~\cite{Feichtenhofer2017DetectTT}  & 50.1 & 24.9 & 23.1 & 27.1 & 3561 & 52481 & 2715 \\ 
        Tracktor-no-FPN & 57.4 & 58.7 & 30.2 & 22.5 & 2821 & 45042 & 1981 \\ 
        \midrule
        Tracktor & 61.5 & 61.1 & 33.5 & \textbf{20.7} & 367 & 42903 & 1747 \\ 
        
        Tracktor+reID & 61.5 & 62.8 & 33.5 & \textbf{20.7} & 367 & 42903 & 921 \\ 

        Tracktor+CMC & \textbf{61.9} & 64.1 & \textbf{35.3} & 21.4 & \textbf{323} & \textbf{42454} & 458 \\ 

        Tracktor++ (reID + CMC) & \textbf{61.9} & \textbf{64.7} & \textbf{35.3} & 21.4 & \textbf{323} & \textbf{42454} & \textbf{326} \\ 


        
        \bottomrule
    \end{tabular}}

\caption{%
This ablation study illustrates multiple aspects on the performance of our Tracktor.
In particular the improvements from extending it with tracking specific methods, i.e., a short-term bounding box re-identification and camera motion compensation by frame alignment.
The combination yields the Tracktor++ tracker.
We evaluated only on the Faster R-CNN set of MOT17 public detections.
%
The arrows indicate low or high optimal metric values.}
\vspace{-0.2cm}
\label{tab:ablation}
\end{table}

The ablation study on the MOT17~\cite{milanarxiv2016} training set in Table~\ref{tab:ablation} is intended to show three aspects: (i) the superiority of our approach when applying a detector for tracking, (ii) the potential from an improved object detection method and (iii) improvements from extending our vanilla Tracktor with tracking specific methods, namely, re-identification (reID) and camera motion compensation (CMC).
%
%
%
It should be noted, that although MOT17Det and MOT17 contain the same images, we refrained from a cross-validation on the training set as our vanilla Tracktor was never trained on tracking ground truth data.
The video object detector and tracker {\it D\&T}~\cite{Feichtenhofer2017DetectTT} trains a detector on tracking ground truth data which generates two-frame tracklets.
However, despite a subsequent offline dynamic programming track generation their detector-based tracker is inferior to our online regression-based track generation over multiple frames.
In addition, we demonstrate the potential of our framework with respect to improved detection methods by showing the tracking performance of {\it Tracktor-no-FPN}, i.e., our approach and a Faster R-CNN without Feature Pyramid Networks (FPN)~\cite{FPN_2017_CVPR}.
Despite the simple nature of our extensions to Tracktor++, their contribution is significant towards the drastic reduction of identity switches and an increment of the IDF1 measure.
%
%
%
%
In the next section, we show that this effect successfully translates to a comparison with other state-of-the-art methods on the test set.
%
%
%
%

\subsection{Benchmark evaluation}


\begin{table}
\center
\tabcolsep=0.11cm

    \resizebox{\columnwidth}{!}{
    \begin{tabular}{c l c c c c c c c c}
     \toprule
          & Method & MOTA $\uparrow$ & IDF1 $\uparrow$ & MT $\uparrow$ & ML $\downarrow$ & FP $\downarrow$ & FN $\downarrow$ & ID Sw. $\downarrow$ \\ [0.5ex] 
     \midrule
     
     \parbox[t]{3mm}{\multirow{6}{*}{\rotatebox[origin=c]{90}{MOT17}}} & Tracktor++ & \textbf{53.5} & 52.3 & 19.5 & 36.6 & \textbf{12201} & 248047 & 2072\\
          & eHAF~\cite{sheng2018} & 51.8 & \textbf{54.7} & \textbf{23.4} & 37.9 & 33212 & \textbf{236772} & 1834 \\
          & FWT~\cite{HenschelLCR17} & 51.3 & 47.6 & 21.4 & 35.2 & 24101 & 247921 & 2648 \\
          & jCC~\cite{jCCpami2018} & 51.2 & 54.5   & 20.9 & 37.0 & 25937 & 247822 & \textbf{1802} \\
          & MOTDT17~\cite{ChenAZS18} & 50.9 & 52.7 & 17.5 & 35.7 & 24069 & 250768 & 2474 \\
          & MHT\_DAM~\cite{KimLCR15} & 50.7 & 47.2 & 20.8 & 36.9 & 22875 & 252889 & 2314 \\
     \midrule
     
     \parbox[t]{3mm}{\multirow{6}{*}{\rotatebox[origin=c]{90}{MOT16}}} & Tracktor++ & \textbf{54.4} & \textbf{52.5} & 19.0 & \textbf{36.9} & \textbf{3280} & \textbf{79149} & 682 \\
          & HCC~\cite{maACCV2019} & 49.3 & 50.7 & 17.8 & 39.9 & 5333 & 86795 & \textbf{391} \\
          & LMP~\cite{TangAAS17} & 48.8 & 51.3 & 18.2 & 40.1 & 6654 & 86245 & 481 \\
          & GCRA~\cite{MaYYZZJX18} & 48.2 & 48.6 & 12.9 & 41.1 & 5104 & 88586 & 821 \\
          & FWT~\cite{HenschelLCR17} & 47.8 & 44.3 & \textbf{19.1} & 38.2 & 8886 & 85487 & 852 \\
          & MOTDT~\cite{ChenAZS18} & 47.6 & 50.9 & 15.2 & 38.3 & 9253 & 85431 & 792 \\ 
     \midrule
     
     \parbox[t]{3mm}{\multirow{5}{*}{\rotatebox[origin=c]{90}{2D MOT 2015}}} & Tracktor++ & \textbf{44.1} & 46.7 & 18.0 & \textbf{26.2} & 6477 & \textbf{26577} & 1318 \\
          & AP\_HWDPL\_p~\cite{ChenASZB17} & 38.5 & \textbf{47.1} & 8.7 & 37.4 & \textbf{4005} & 33203 & 586 \\
          & AMIR15~\cite{SadeghianAS17} & 37.6 & 46.0 & 15.8 & 26.8 & 7933 & 29397 & 1026 \\
          & JointMC~\cite{jCCpami2018} & 35.6 & 45.1 & \textbf{23.2} & 39.3 & 10580 & 28508 & 457 \\
          & RAR15pub~\cite{FangXS17} & 35.1 & 45.4 & 13.0 & 42.3 & 6771 & 32717 & \textbf{381} \\
     \bottomrule
    \end{tabular}}

\caption{We compare our online multi-object tracker Tracktor++ with other modern tracking methods. As a result, we achieve a new state-of-the-art in terms of MOTA for public detections on all three MOTChallenge benchmarks. The arrows indicate low or high optimal metric values.}
\vspace{-0.2cm}
\label{tab:mot}

\end{table}

We evaluate the performance of our Tracktor++ on the test set of the respective benchmark, without any training or optimization on the tracking train set.
Table~\ref{tab:mot} presents the overall results accumulated over all sequences, and for MOT17 over all three sets of public detections.
For our comparison, we only consider officially published and peer-reviewed entries in the MOTChallenge benchmark.
Our supplementary material provides a detailed summary of all results on individual sequences.
For all sequences, camera motion compensation (CMC) and reID are used.
The only low frame rate sequence is the 2D MOT 2015 {\it AVG-TownCentre}, for which we apply the aforementioned constant velocity assumption (CVA).  
%
For the two autonomous driving sequences, originally from the KITTI~\cite{kittiGeiger2012CVPR} benchmark, we apply the rotation as well as translation camera motion compensation. 
Note, we use the same Tracktor++ tracker, trained on MOT17Det object detections, for all benchmarks.
As we show, it is able to achieve a new  state-of-the-art in terms of MOTA on all three challenges.
%

In particular, our results on MOT16 demonstrate the ability of our tracker to cope with detections of comparatively minor performance.
Due to the nature of our tracker and the robustness of the frame by frame bounding box regression, we outperform all other trackers on MOT16 by a large margin, specifically in terms of false negatives (FN) and identity preserving (IDF1).
It should be noted, that we also provide a new state-of-the-art on 2D MOT 2015, even though the characteristics of the scenes are very different from MOT17. We do not use MOT15 training sequences, which further illustrates the generalization strength of our tracker.

\section{Analysis}
\label{sec:analysis}


\begin{figure}
    \centering
    \includegraphics[width=\columnwidth]{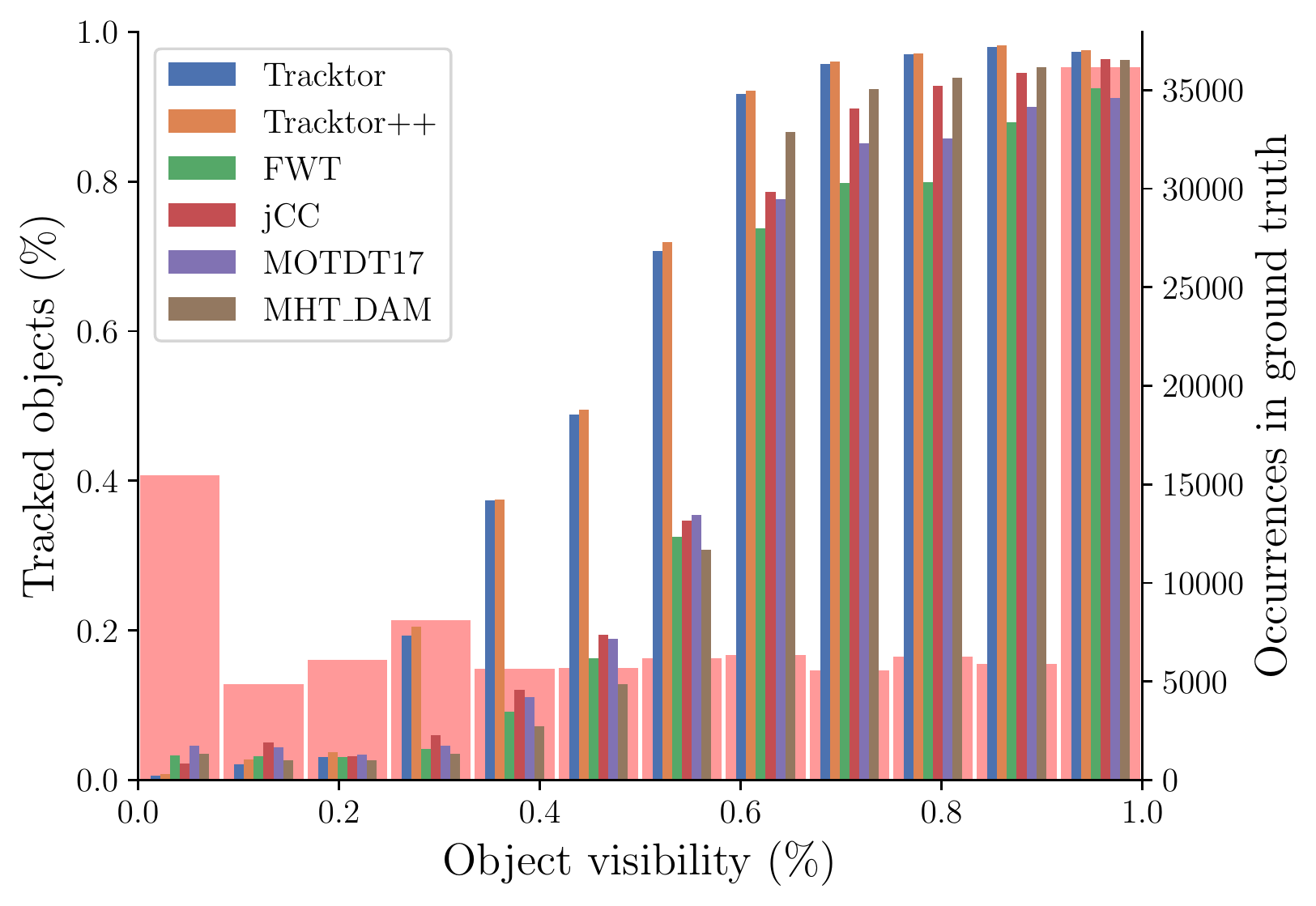}
    \caption{We illustrate the ratio of tracked objects with respect to their visibility evaluated on the Faster R-CNN public detections. The results clearly demonstrate that none of the presented more sophisticated methods achieves superior performance to our approach. This is especially noticeable for highly occluded boxes. The transparent red bars indicate the ground truth distribution of visibilities. }
    \vspace{-0.2cm}
    \label{fig:vis}
\end{figure}

\begin{figure*}[t!]
    \centering

    \subfloat[DPM detections]{
        \label{fig:heights09-DPM}
        \setlength\tabcolsep{0pt}
        \begin{tabular}{c}
            \includegraphics[width=0.33\textwidth]{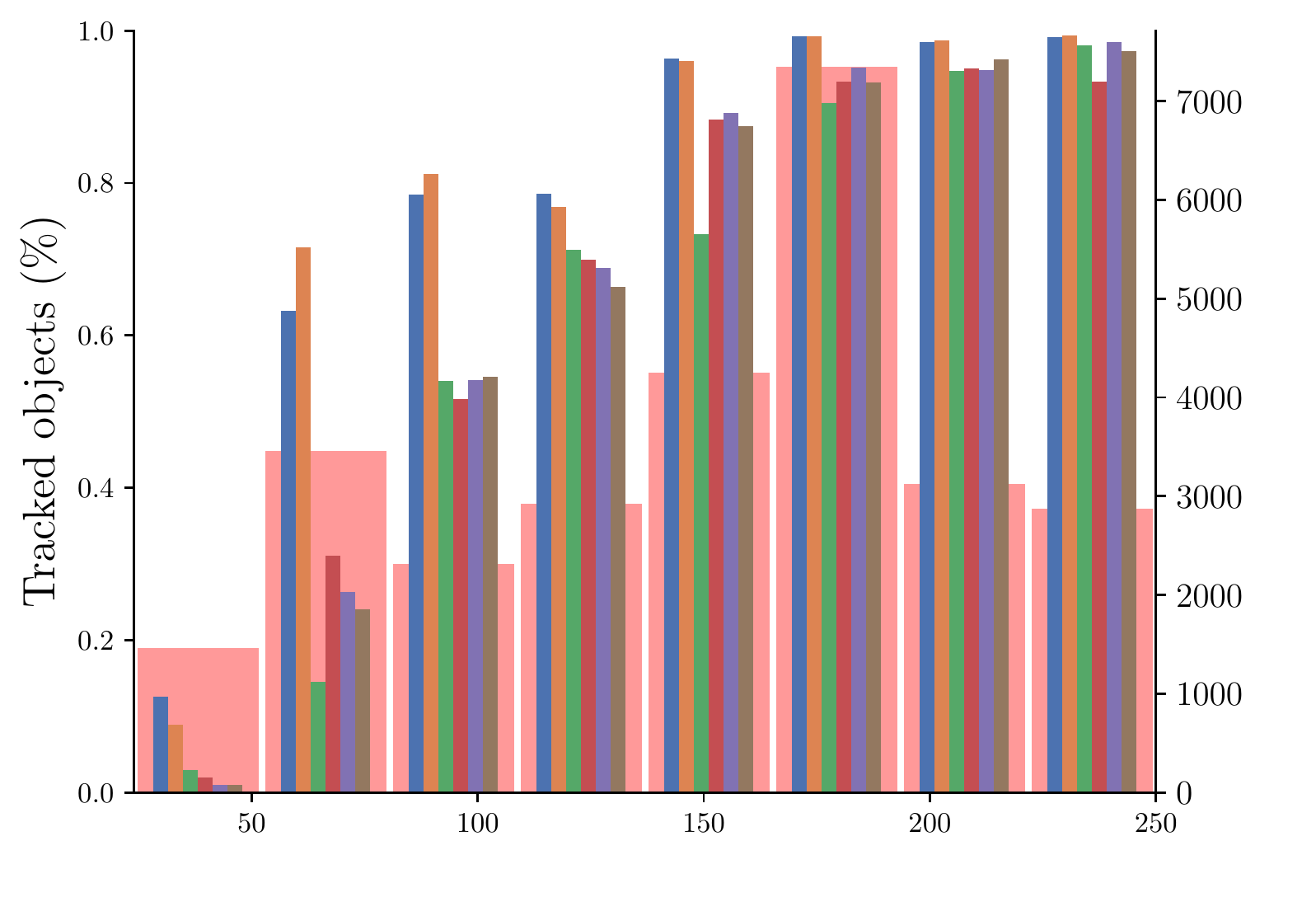}
            \vspace{-0.2cm} \\
            \includegraphics[width=0.33\textwidth]{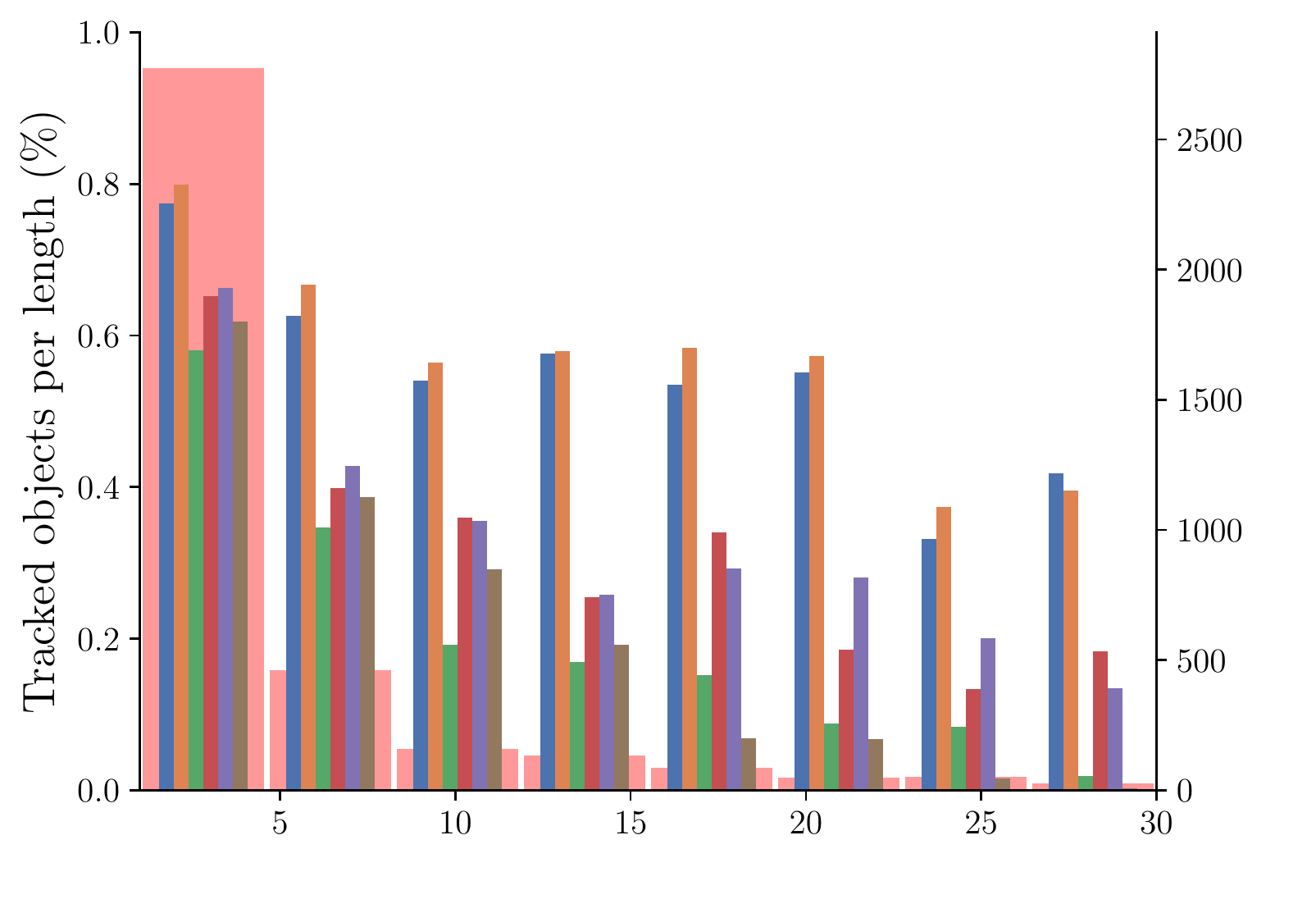}
        \end{tabular}}
    \subfloat[Faster R-CNN detections]{
        \label{fig:heights09-FRCNN}
        \setlength\tabcolsep{0pt}
        \begin{tabular}{c}
            \includegraphics[width=0.33\textwidth]{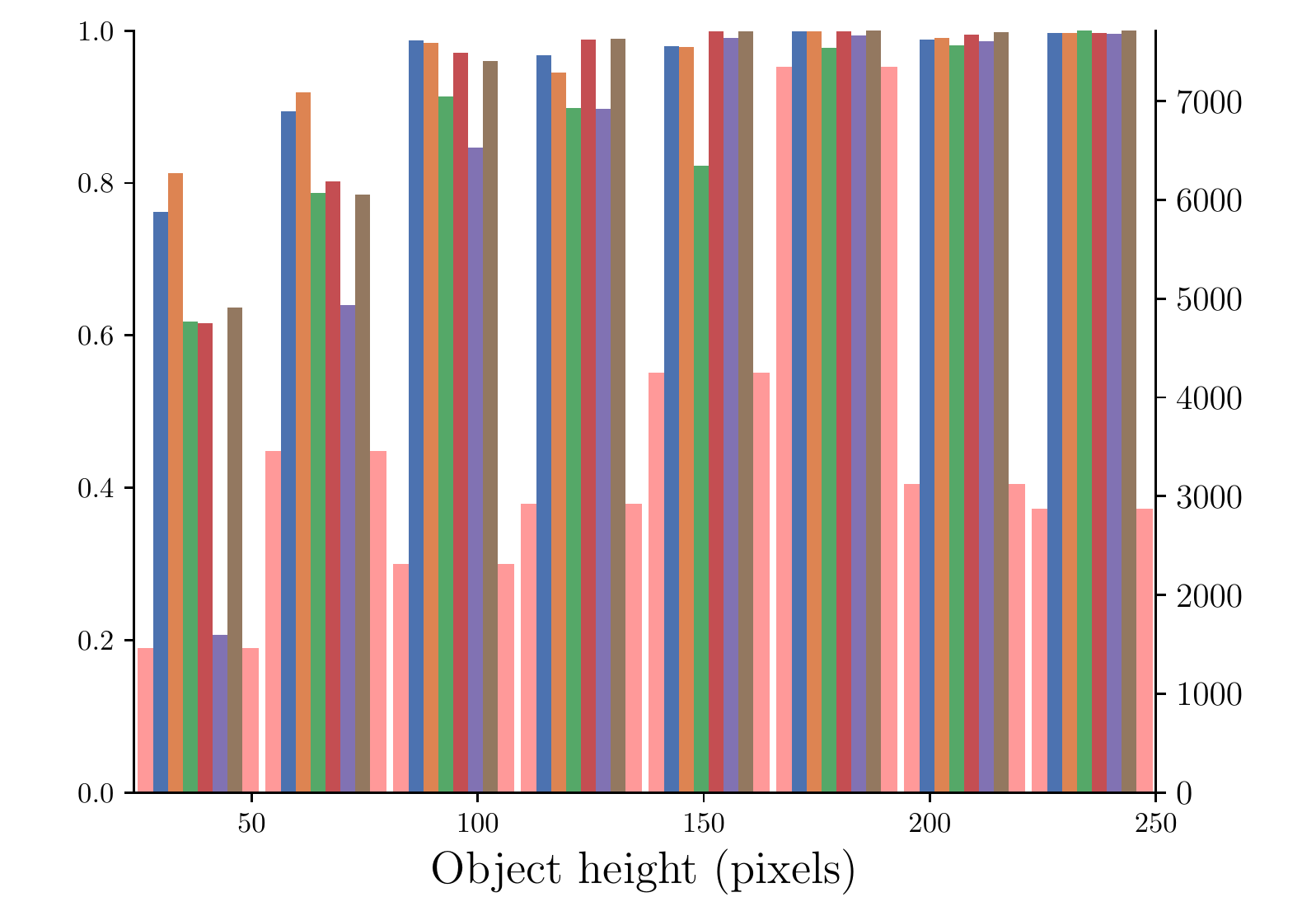}
            \vspace{-0.2cm} \\
            \includegraphics[width=0.33\textwidth]{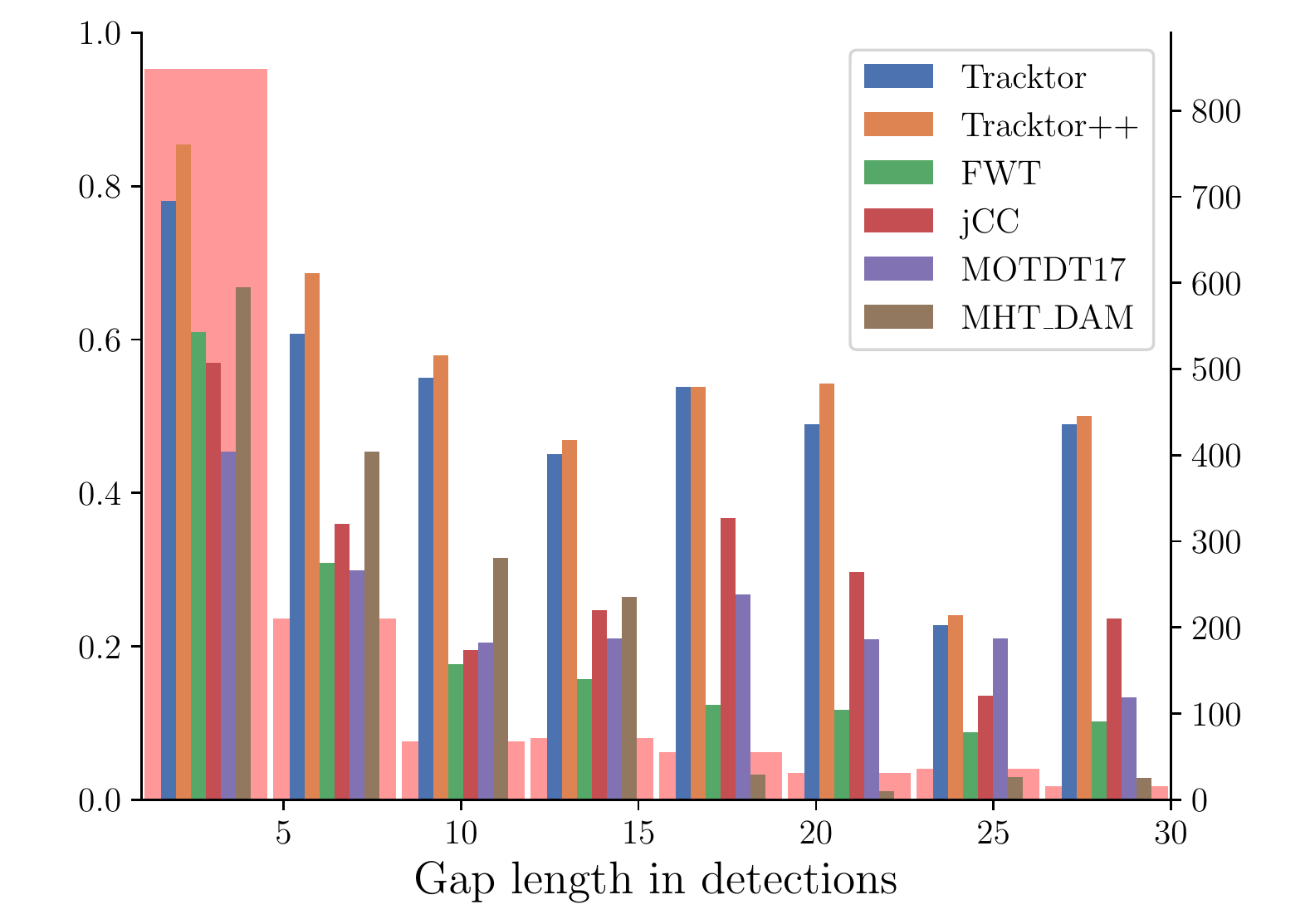}
        \end{tabular}}
    \subfloat[SDP detections]{
        \label{fig:heights09-SDP}
        \setlength\tabcolsep{0pt}
        \begin{tabular}{c}
            \includegraphics[width=0.33\textwidth]{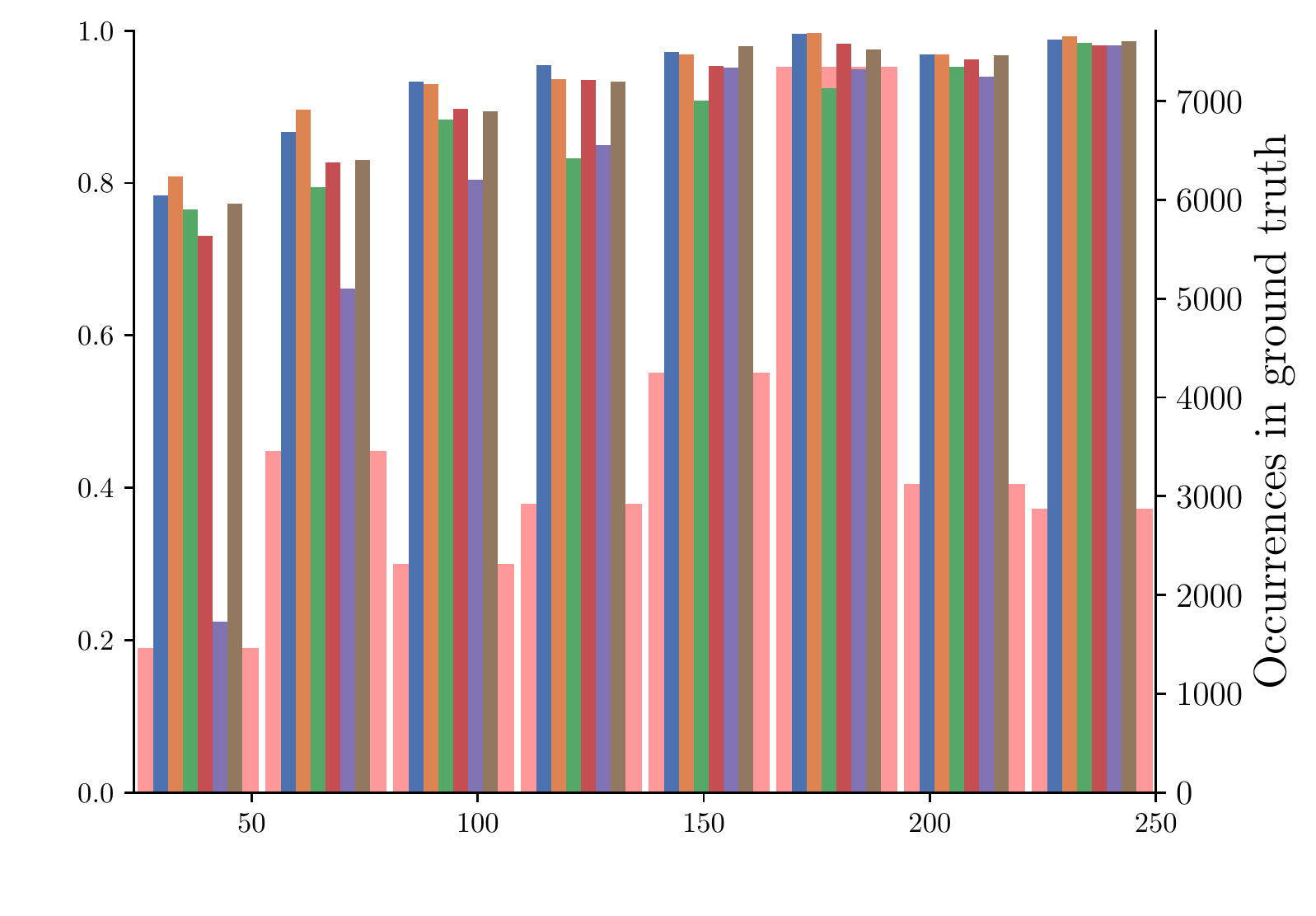}
            \vspace{-0.2cm} \\
            \includegraphics[width=0.33\textwidth]{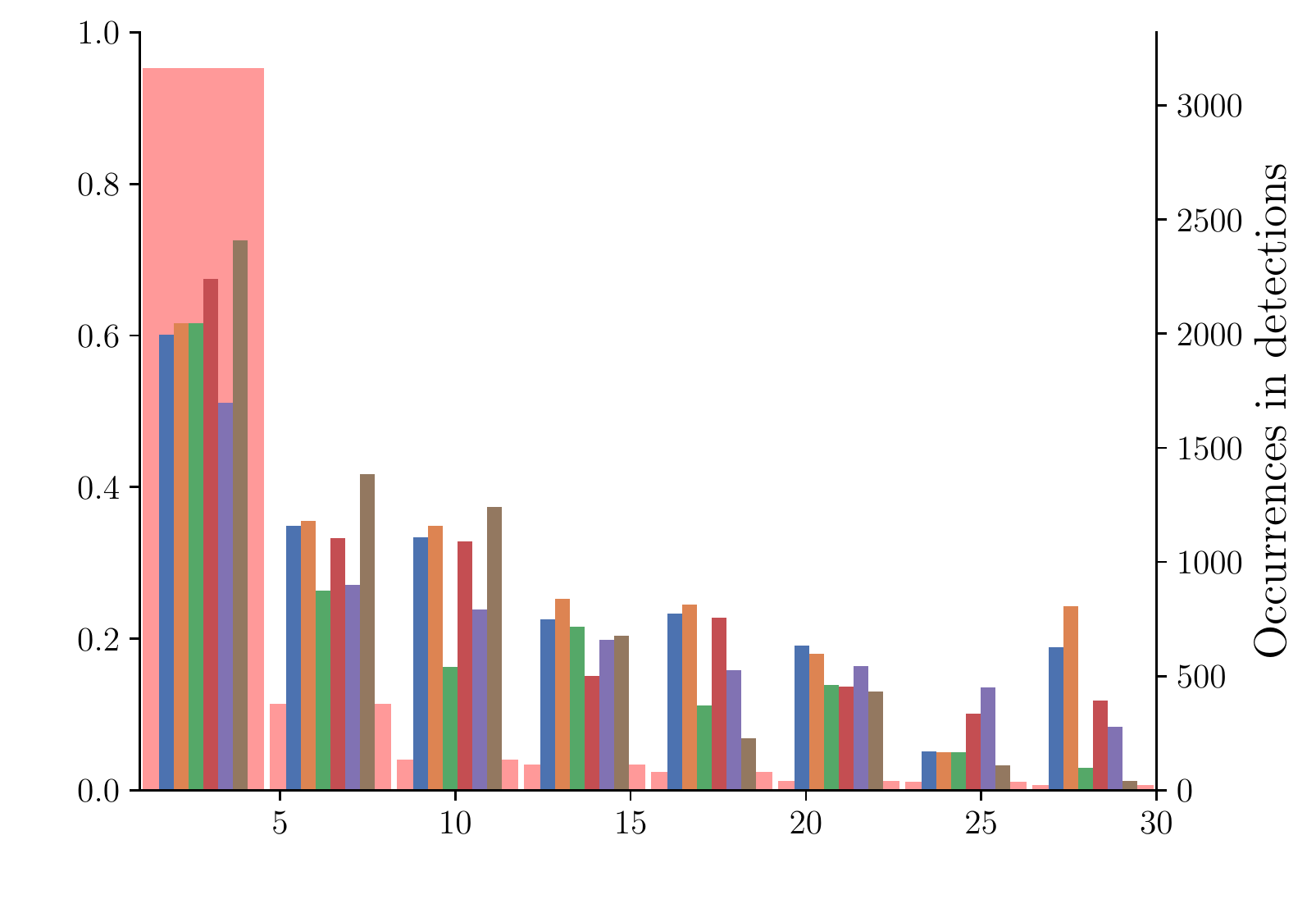}
        \end{tabular}}
    
    \caption{
    The two rows illustrate the ratio of tracked objects with respect to: (i) object heights and (ii) the length of gaps in the provided public detections.
    The transparent red bars indicate the ground truth distribution of heights and gap lengths in the detections, respectively.
    To demonstrate the shortcomings of the presented trackers we limited the height comparison to objects with visibility greater or equal than 0.9.
    Tracks that are not detected at all are not considered as a gap.
    Hence, SPD generates the most gaps.
    For it also provides the most detections.}
    \label{fig:heights09}
\end{figure*}

\begin{table}
\center 
\tabcolsep=0.11cm

    \resizebox{\columnwidth}{!}{
    \begin{tabular}{l c c c c c c}
        \toprule
        Method                      & Online    & Graph & reID     & Appearance model & Motion model & Other \\ [0.5ex] 
        \midrule
        Tracktor                    & $\times$ &        &          &          &               &                       \\
        Tracktor++                  & $\times$ &        & $\times$ &          & Camera        &                       \\
        FWT~\cite{HenschelLCR17}    &          & Dense  &          &          &               & Face detection        \\
        jCC~\cite{jCCpami2018}      &          & Dense  &          &          &               & Point trajectories    \\
        MOTDT17~\cite{ChenAZS18}    & $\times$ &        & $\times$ & $\times$ & Kalman        &                       \\
        MHT\_DAM~\cite{KimLCR15}    &          & Sparse &          & $\times$ & Kalman        &                       \\
        \bottomrule
    \end{tabular}}

\caption{A summary of the fundamental characteristics of our methods and other state-of-the-art trackers.}
\vspace{-0.2cm}
\label{tab:trackers}
\end{table}

The superior performance of our tracker without any tracking specific training or optimization demands a more thorough analysis.
Without sophisticated tracking methods, it is not expected to excel in crowded and occluded, but rather only in benevolent, tracking scenarios.
Which begs the question whether more common tracking methods fail to specifically address these complex scenarios as well.
Our experiments and the subsequent analysis ought to demonstrate the strengths of our approach for easy tracking scenarios and motivate future research to focus on remaining complex tracking problems.
%
%
In particular, we question the common execution of tracking-by-detection and suggest a new tracking paradigm.
%
The subsequent analysis is conducted on the MOT17 training data and we compare all top performing methods with publicly shared data.

\subsection{Tracking challenges}

For a better understanding of our tracker, we want to analyse challenging tracking scenarios and compare its strengths and weaknesses to other trackers.
To this end, we summarize their fundamental characteristics in Table~\ref{tab:trackers}.
FWT~\cite{HenschelLCR17} and jCC~\cite{jCCpami2018} both apply a dense offline graph optimization on all detections in a given sequence.
In contrast, MHT\_DAM~\cite{KimLCR15} limits its optimization to a sparse forward view of hypothetical trajectories.

\noindent{\bf Object visibility.}
Intuitively, we expect diminished tracking performance for object-object or object-non-object occlusions, i.e., for targets with diminished visibility.
In Figure~\ref{fig:vis}, we compare the ratio of successfully tracked bounding boxes with respect to their visibility.
%
%
The transparent red bar indicates the occurrences of ground truth bounding boxes for each visibility, and illustrates the proportionate impact on the overall performance of the trackers.
Our method achieves superior performance even for partially occluded bounding boxes with visibilities as low as 0.3.
%
Neither the identify preserving aspects of MHT\_DAM and MOTDT17~\cite{ChenAZS18} nor the offline interpolation capabilities of MHT\_DAM and jCC seem to successfully tackle highly occluded objects.
%
%
The high MOTA values in Table~\ref{tab:mot} are largely due to the unbalanced distribution of ground truth visibilities.
%
%
%
As expected, our extended version only achieves minor improvements over our vanilla Tracktor.

\noindent{\bf Object size.}
%
%
In view of the large fraction of visible but not tracked objects in Figure~\ref{fig:vis}, we argue that the trackability of an object is not only dependent on its visibility, but also its size.
Therefore, we conduct the same comparison as for the visibility but for the size of an object.
In the first row of Figure~\ref{fig:heights09}, we assume the height of a pedestrian to be proportional to its size and compare on all three MOT17 public detection sets.
%
%
All methods performed similarly well for object heights larger than 250 pixels.
To demonstrate their shortcomings even for highly visible objects, we only compare objects with a visibility larger than 0.9.
As expected, the trackability of an object decreases drastically with its size across all three detection sets.
%
%
%
%
Our tracker shows its strength in compensating for insufficient DPM and Faster R-CNN detections for all object sizes.
All methods except MOTDT17 benefit from the additional small detections provided by SDP.
For our tracker this is largely due to the Feature Pyramid Network extension of our Faster-RCNN detector.
However, the learned appearance model and reID of the online MOTDT17 method seem generally vulnerable to small detections.
Appearance models generally suffer from small object sizes and few observed pixels.
%
%
%
In conclusion, except from our compensation of inferior detections none of the trackers exhibit a notably better performance with respect to varying object sizes.

\noindent{\bf Robustness to detections.}
%
%
The performance of tracking-by-detection methods with respect to visibility and size is inherently limited by the robustness of the underlying detection method.
However, as observed for the object size, trackers differ in their ability to cope with, or benefit from, varying quality of detections.
%
In the second row of Figure~\ref{fig:heights09}, we quantify this ability in terms of detection gaps on their coverage by the tracker. 
%
%
We define a detection gap as part of a ground truth trajectory that was at least once detected, and compare coverage of each gap vs. the gap length.
%
%
Intuitively, long gaps are harder to compensate for, as the online or offline tracker has to perform a longer hallucination or interpolation, respectively.
We indicated the occurrences of gap lengths over the respective set of detections in transparent red.
For DPM and Faster R-CNN detections, two solutions lead to notable gap coverage: (i) offline interpolation such as in jCC, or (ii) motion prediction with Kalman filter and reID as in MOTDT.
Compared to the graph-based jCC method, the online MOTDT17 method excels at covering particularly long gaps.
%
%
However, none of these dedicated tracking methods yields similar robustness to our frame by frame regression tracker, which achieves far superior coverage.
This holds especially true for long detection gaps with more than 15 frames.
%
%
Offline methods benefit the most from improved SDP detections and neither our nor the MOTDT17 tracker convince with a notable gap length robustness.
%
%

%
\noindent{\bf Identity preservation.} 
The results of our Tracktor++ summarized in Table~\ref{tab:mot} indicate an identity preservation performance in terms of IDF1 and identity switches comparable with dedicated tracking methods.
This is achieved without any offline graph optimization as in jCC~\cite{jCCpami2018} or eHAF~\cite{sheng2018}.
In particular, MOTDT17, which applies a sophisticated appearance model and reID, is not substantially superior to our regression tracker and its comparatively simple extensions.
However, our method excels in reducing the number of false positives in MOT17 as well as MOT16.
%
%
%
In addition, we have shown that our Tracktor is capable of incorporating additional identity preserving extension.

\subsection{Oracle trackers}


\begin{table}
\center 
\tabcolsep=0.11cm

    \resizebox{\columnwidth}{!}{
    \begin{tabular}{l c c c c c}
        \toprule
        Method & MOTA $\uparrow$ & IDF1 $\uparrow$ & FP $\downarrow$ & FN $\downarrow$ & ID Sw. $\downarrow$ \\ [0.5ex] 
        \midrule


        Tracktor & 61.5 & 61.1 & 367 & 42903 & 1747 \\ 
        Tracktor++ & \textcolor{LimeGreen}{+0.4} & \textcolor{LimeGreen}{+3.6} & \textcolor{LimeGreen}{-44} & \textcolor{LimeGreen}{-449} & \textcolor{LimeGreen}{-1421} \\ 
        \hline
        
        Oracle-Kill & \textcolor{LimeGreen}{+0.7} & \textcolor{Red}{-0.7} & \textcolor{LimeGreen}{-178} & \textcolor{LimeGreen}{-694} & \textcolor{Red}{+129} \\ 

        Oracle-REG & \textcolor{LimeGreen}{+1.4} & \textcolor{LimeGreen}{+5.6} & \textcolor{LimeGreen}{-218} & \textcolor{LimeGreen}{-1401} & \textcolor{LimeGreen}{-1463} \\ 
        \hline
        Oracle-MM & \textcolor{LimeGreen}{+0.9} & \textcolor{LimeGreen}{+5.2} & \textcolor{LimeGreen}{-168} & \textcolor{LimeGreen}{-898} & \textcolor{LimeGreen}{-1332} \\ 
        
        Oracle-reID & 0.0 & \textcolor{LimeGreen}{+10.0} & 0 & 0 & \textcolor{LimeGreen}{-1094} \\ 
        
        Oracle-MM-reID & \textcolor{LimeGreen}{+0.9} & \textcolor{LimeGreen}{+13.9} & \textcolor{LimeGreen}{-168} & \textcolor{LimeGreen}{-898} & \textcolor{LimeGreen}{-1706} \\ 
        
        
        Oracle-MM-reID-INTER & \textcolor{LimeGreen}{+2.6} & \textcolor{LimeGreen}{+15.9} & \textcolor{Red}{+3774} & \textcolor{LimeGreen}{-6769} & \textcolor{LimeGreen}{-1680} \\ 
        \hline

        Oracle-ALL & \textcolor{LimeGreen}{+10.7} & \textcolor{LimeGreen}{+22.5} & \textcolor{LimeGreen}{-360} & \textcolor{LimeGreen}{-11745} & \textcolor{LimeGreen}{-1743} \\ 
        \bottomrule
    \end{tabular}}
    \vspace{-0.15cm}
\caption{%
To show the potential of Tracktor and indicate promising future research directions, we present multiple oracle trackers.
Each oracle exploits ground truth data for a specific task, simulating, e.g., a perfect re-identification (reID) or motion model (MM).
We evaluate only on the Faster R-CNN set of MOT17 public detections and highlight performance gains and losses with respect to the vanilla Tracktor in green and red, respectively.
%
The arrows indicate low or high optimal metric values.
}
\vspace{-0.4cm}
\label{tab:oracle}
\end{table}

We have shown that none of the dedicated tracking methods specifically targets challenging tracking scenarios, i.e., objects under heavy occlusions or small objects.
We therefore want to motivate our Tracktor as a new tracking paradigm.
To this end, we analyse our performance two-fold: (i) the impact of the object detector on the killing policy and bounding box regression, (ii) identify performance upper bounds for potential extensions to our Tracktor.
In Table~\ref{tab:oracle}, we present several oracle trackers by replacing parts of our algorithm with ground truth information.
If not mentioned otherwise, all other tracking aspects are handled by our vanilla Tracktor.
Their analysis should provide researchers with useful insights regarding the most promising research directions and extensions of our Tracktor.

\noindent{\bf Detector oracles.}
To simulate a potentially perfect object detector, we introduce two oracles:
%
%
%
\begin{itemize}
\item Oracle-Kill: Instead of killing with NMS or classification score we use ground truth information.
\item Oracle-REG: Instead of regression, we place the bounding boxes at their ground truth position.
\end{itemize}
%
%
%
Both oracles yield substantial improvements with respect to MOTA and FP.
However, killing by ground truth instead of score deteriorates identity preservation as the regression struggles with otherwise unseen bounding boxes.


%
\noindent{\bf Extension oracles.}
It should be noted, that Tracktor++ with non-perfect extensions already compensates for some of the detector's insufficiencies.
%
The reID and motion model (MM) oracles simulate potential additional performance gains.
In order to remain online, these exclude any form of hindsight tracking-gap interpolation.
%
\begin{itemize} 
\item Oracle-MM: A motion model places each bounding box at the center of the ground truth in the next frame. 
\item Oracle-reID: Re-identification is performed with ground truth identities.
\end{itemize}
As expected, both oracles improve IDF1 and identity switches substantially.
The combined Oracle-MM-reID represents the extension upper bound of Tracktor++.
%
%
%

\noindent{\bf Omniscient oracle.}
Oracle-ALL performs ground truth killing, regression and reID.
%
%
We consider its top MOTA of 72.2\%, in combination with a high IDF1 and virtually no false positives, as the absolute upper bound of Tracktor with a Faster R-CNN and FPN object detector.
%



%

The substantial performance gains from Oracle-MM indicate the potential of extending Tracktor with a sophisticated motion model.
In particular, Oracle-MM-reID-INTER suggests a predictive motion model which hallucinates the position of an object through long occlusions.
Such a motion model avoids offline post processing and additional false positives from wrong linear occlusion paths caused by long detection gaps and camera movement

%


%

\subsection{Towards a new tracking paradigm}
To conclude our analysis we propose two approaches on how to utilize Tracktor as a starting point for future research directions:

\noindent{\bf Tracktor with extensions.}
Apply Tracktor to a given set of detections and extend it with tracking specific methods.
Scenarios with large and highly visible objects will be covered by the frame to frame bounding box regression.
%
%
For the remaining, it seems most promising to implement a hallucinating motion model, taking into account the individual movements of objects.
In addition, such a motion predictor reduces the necessity for an advanced killing policy.
 
%
%
%

\noindent{\bf Tracklet generation.}
Analogous to tracking-by-detection, we propose a tracking-by-tracklet approach.
Indeed, many algorithms already use tracklets as input \cite{henschelgcpr2014,zamireccv2012}, as they are richer in information for computing motion or appearance models.
However, usually a specific tracking method is used to create these tracklets.
We advocate the exploitation of the detector itself, not only to create sparse detections, but frame to frame tracklets.
The remaining complex tracking cases ought to be tackled by a subsequent tracking method. 

In this work, we have formally defined those hard cases, analyzing the situations in which not only our method but other dedicated tracking solutions fail.
And by doing so, we question the current focus of research in multi-object tracking, in particular, the missing confrontation with challenging tracking scenarios.



\section{Conclusions}

We have shown that the bounding box regressor of a trained Faster-RCNN detector is enough to solve most tracking scenarios present in current benchmarks.
A detector converted to Tracktor needs no specific training on tracking ground truth data and is able to work in an online fashion.
In addition, we have shown that our Tracktor is extendable with re-identification and camera motion compensation, providing a substantial new state-of-the-art on the MOTChallenge.
%
%
%
We analyzed the performance of multiple dedicated tracking methods on challenging tracking scenarios and none yielded substantially better performance compared to our regression based Tracktor.
We hope this work establishes a new tracking paradigm, utilizing the object detector's full capabilities.

\small
\PAR{Acknowledgements.} This research was funded by the Humboldt Foundation through the Sofja Kovalevskaja Award.

\ifarxiv
    \pagenumbering{gobble}
    \setcounter{section}{0}
    \setcounter{figure}{0}
    \setcounter{table}{0}
    \setcounter{equation}{0}
    \setcounter{footnote}{0}
    \def\paperTitle{Tracking without bells and whistles \\ {\normalfont Supplementary Material}}
\unless\ifarxiv
    \title{\paperTitle}


\author{\qquad Philipp Bergmann%
\thanks{Contributed equally. Correspondence to: tim.meinhardt@tum.de}\qquad
\and \qquad Tim Meinhardt
\footnotemark[1]\qquad
\and \qquad Laura Leal-Taixe\qquad
\vspace{0.2cm}
\and
\qquad \normalfont{\textit{Technical University of Munich}}\qquad
}


\maketitle
\ifarxiv
    \thispagestyle{fancy}
\fi
\fi
\ifarxiv
    \title{\paperTitle}

    \author{\qquad Philipp Bergmann%
    \footnotemark[1]\qquad
    \and \qquad Tim Meinhardt
    \footnotemark[1]\qquad
    \and \qquad Laura Leal-Taixe\qquad
    \vspace{0.2cm}
    \and
    \qquad \normalfont{\textit{Technical University of Munich}}\qquad
    }

    \maketitle
\fi

\thispagestyle{empty}
\ifarxiv
    \thispagestyle{fancy}
\fi

\begin{abstract}
The supplementary material complements our work with the pseudocode representation of Tracktor and additional implementation and training details of its object detector and tracking extensions.
In addition, we provide more details on our experiments and analysis including the MOTChallenge benchmark results of our Tracktor++ tracker for each sequence and set of public detections.
%
\end{abstract}

\appendix
\section{Implementation}

For the sake of completeness and in order to facilitate the reproduction of our results, we provide additional implementation details and references of our Tracktor and its extensions.

\subsection{Tracktor}

In Algorithm~\ref{alg:method_tracktor_private} and~\ref{alg:method_tracktor_public}, we present a structured pseudocode representation of our Tracktor for private and public detections, respectively.
Algorithm~\ref{alg:method_tracktor_private} corresponds to the method illustrated in Figure~\ref{fig:method_vis} and Section~\ref{sec:tracktor} of our main work.
%

\paragraph{Object detector.}
As mentioned before, our approach requires no dedicated training or optimization on tracking ground truth data and performs tracking only with an object detection method.
To this end, we train the Faster R-CNN (FRCNN)~\cite{rennips2015} multi-object detector with Feature Pyramid Networks (FPN)~\cite{FPN_2017_CVPR} on the MOT17Det~\cite{milanarxiv2016} dataset.

In addition, we follow the improvements suggested by~\cite{ChenG17a}.
These include a replacement of the Region of Interest (RoI) pooling~\cite{Girshick2015FastR} by the {\it crop and resize} pooling suggested by Huang et al.~\cite{HuangRSZKFFWSG016} and training with a batch size of $N=1$ instead of $N=2$ while increasing the number of extracted regions from $R=128$ to $R=256$. 
These changes and the addition of FPN ought to improve the detection results for comparatively small objects.
%
%
We achieve the best results with a ResNet-101~\cite{HeZRS15} as the underlying feature extractor.
In Table~\ref{tab:detectors}, we compare the performance on the official MOT17Det detection benchmark for the three object detection methods mentioned in this work.
The results demonstrate the incremental gain in detection performance of DPM~\cite{dpmpami2009}, FRCNN and SDP~\cite{sdpYangcvpr2016} (ascending order).
Our FRCNN implementation without FPN is on par with the official MOT17Det entry and represents the detector applied in the {\it Tracktor-no-FPN} variant of our ablation study in Section~\ref{sec:ablation}.
%


\begin{table}[h]
\center 
\tabcolsep=0.11cm

    \resizebox{\columnwidth}{!}{
    \begin{tabular}{l c c c c c c}
        \toprule
        Method & AP $\uparrow$ & MODA $\uparrow$ & FP $\downarrow$ & FN $\downarrow$ & Precision $\uparrow$ & Recall $\uparrow$ \\ [0.5ex] 
        \midrule  
        FRCNN + FPN                         & 0.81 & 70.2 & 14914 & 19196 & \textbf{96.5} & 83.2 \\
        FRCNN                               & 0.72 & 71.6 & 8227 & 24269 & 91.6 & 78.8 \\
        \midrule
        DPM~\cite{dpmpami2009}              & 0.61 & 31.2 & 42308 & 36557 & 64.8 & 68.1 \\
        FRCNN~\cite{rennips2015}            & 0.72 & 68.5 & 10081 & 25963 & 89.8 & 77.3 \\
        SDP~\cite{sdpYangcvpr2016}          & \textbf{0.81} & \textbf{76.9} & \textbf{7599} & \textbf{18865} & 92.6 & \textbf{83.5} \\
        \bottomrule
    \end{tabular}}

\caption{A comparison of our Faster R-CNN (FRCNN) with Feature Pyramid Networks (FPN) implementation on the MOT17Det detection benchmark with the three object detection methods mentioned in this work. Our vanilla FRCNN results are on par with the official FRCNN implementation. The extension with FPN yields a detection performance close to SDP. For a detailed summary of the shown detection metrics we refer to the official MOTChallenge web page:~\url{https://motchallenge.net}.}

\label{tab:detectors}
\end{table}

\subsection{Tracking extensions}

Our presented Tracktor++ tracker is an extension of the Tracktor that uses two multi-pedestrian tracking specific extensions, namely, a motion model and re-identification.

\paragraph{Motion model.}
For the motion model via camera motion compensation (CMC) we apply image registration using the Enhanced Correlation Coefficient (ECC) maximization as in \cite{EvangelidisP08}.
The underlying image registration allows either for an euclidean or affine image alignment mode.
We apply the first for rotating camera movements, e.g., as a result of an unsteady camera movement.
In the case of an additional camera translation such as in the autonomous driving sequences of 2D MOT 2015~\cite{milanarxiv2016}, we resort to the affine transformation.
It should be noted that in MOT17~\cite{milanarxiv2016}, camera translation is comparatively slow and therefore we consider all sequences as only rotating.
In addition, we present a second motion model which aims at facilitating the regression for sequences with low frame rates, i.e., large object displacements between frames.
Before we perform bounding box regression, the constant velocity assumption (CVM) model shifts bounding boxes in the direction of their previous velocity.
This is achieved by moving the center of the bounding box ${\bf b}^k_{t-1}$ by the vectorial difference of the two previous bounding box centers at $t-2$ and $t-1$.
The CVA motion model is only applied to the {\it AVG-TownCentre} sequence of 2D MOT 2015.

\paragraph{Re-identification.}
Our short-term re-identification utilizes a Siamese neural network to compare bounding box features and return a measure of their identity.
To this end, we train the TriNet~\cite{HermansBL17} architecture which is based on ResNet-50~\cite{HeZRS15} with the triplet loss and {\it batch hard} strategy as presented in~\cite{HermansBL17}.
The network is optimized with Adam~\cite{adam2014} with $\beta=(0.9,0.999)$ and a decaying learning rate as described in~\cite{HermansBL17}.
Training samples with corresponding identity are generated from the MOT17 tracking ground truth training data.
The TriNet architecture requires input data with a dimension of $H \times W = 256 \times 128$.
To allow for a subsequent data augmentation via horizontal flip and random cropping, each ground truth bounding box is cropped and resized to $\frac{9}{8}(H \times W)$.
A training batch consists of $18$ randomly selected identities, each of which is represented with $4$ different samples.
Identities with less than 4 samples in the ground truth data are discarded.



\section{Experiments}

A detailed summary of our official and published MOTChallenge benchmark results for our Tracktor++ tracker is presented in Table~\ref{tab:tracktorplusplus_verbose}. For the corresponding results for each sequence and set of detections for the other trackers mentioned in this work we refer to the official MOTChallenge web page available at~\url{https://motchallenge.net}.

\subsection{Evaluation metrics}
In order to measure the performance of a tracker, we mentioned the Multiple Object Tracking Accuracy (MOTA)~\cite{clear} and ID F1 Score (IDF1)~\cite{ristanieccvw2016}.
However, previous Tables such as~\ref{tab:tracktorplusplus_verbose} included additional informative metrics.
The false positives (FP) and negatives (FN) account for the total number of either bounding boxes not covering any ground truth bounding box or ground truth bounding boxes not covered by any bounding box, respectively.
To measure the track identity preserving capabilities, we report the total number of identity switches (ID Sw.), i.e., a bounding box covering a ground truth bounding box from a different track than in the previous frame.
The mostly tracked (MT) and mostly lost (ML) metrics provide track wise information on how many ground truth tracks are covered by bounding boxes for either at least 80\% or at most 20\%, respectively.
MOTA and IDF1 are meaningful combinations of the aforementioned basic metrics.
All metrics were computed using the official evaluation code provided by the MOTChallenge benchmark.

\subsection{Raw DPM detections}
As most object detection methods, DPM applies a final non-maximum-suppression (NMS) step to a large set of raw detections.
The MOT16~\cite{milanarxiv2016} benchmark provides both, the set before and after the NMS, as public DPM detections. 
However, this NMS step is performed with DPM classification scores and an unknown Intersection over Union (IoU) threshold.
Therefore, we extracted our own classification scores for all raw detections and applied our own NMS step.
Although not specifically provided, we followed the convention to also process raw DPM detections for MOT17.
Note, several other public trackers already work on raw detections~\cite{jCCpami2018,ChenAZS18,yoonavss2018} and their own classification score and NMS procedure. Therefore, we consider the comparison with public trackers as fair.


\subsection{Evaluation on public detections}
By reclassifying and regressing the given public detections with a private object detector, Tracktor reduces the equalizing effect of public detections to the initialization of new tracks.
In addition to our remarks in Section 3 regarding the {\it publicness} of our method, we emphasize the potential of Tracktor in comparison with other state-of-the-art trackers even without the advantage of the reclassification and regression.
To this end, we show Table~\ref{tab:mot_frcnn_data_association}, which evaluates all trackers on the MOT17 test set only with Faster R-CNN public detections.
Tracktor-no-FPN++ (without Feature Pyramid Networks) uses a vanilla Faster R-CNN for reclassification and regression, effectively, not altering the public detections.
However, the results support the overall conclusions from Table 2 of our main work.


\begin{table}
\center
\tabcolsep=0.11cm

    \resizebox{\columnwidth}{!}{
    \begin{tabular}{l c c c c c c c}
     \toprule
          Method & MOTA $\uparrow$ & IDF1 $\uparrow$ & MT $\uparrow$ & ML $\downarrow$ & FP $\downarrow$ & FN $\downarrow$ & ID Sw. $\downarrow$ \\
     
    \midrule
     

        Tracktor++ & 42.14 & 45.76 & 18.17 & 38.93 & 3918 & 83904 & 648 \\
        \midrule
        Tracktor-no-FPN++ & \textbf{39.41} & 43.46 & 16.63 & 39.00 & \textbf{6975} & \textbf{83380} & 922 \\
        eHAF17 & 37.37 & \textbf{46.44} & \textbf{20.63} & 35.83 & 11050 & 86510 & 605 \\
        FWT & 39.06 & 42.07 & 17.60 & 37.53 & 8397 & 88290 & 780 \\
        jCC & 37.64 & 46.66 & 18.70 & 36.33 & 9984 & 86897 & \textbf{577} \\
        MOTDT17 & 38.81 & 46.34 & 14.47 & 36.91 & 8911 & 88773 & 731 \\
        MHT\_DAM & 37.54 & 46.17 & 17.43 & \textbf{34.86} & 9795 & 89294 & 742 \\

     \bottomrule
    \end{tabular}}

\caption{Comparison on MOT17 test set with Faster R-CNN public detections. Tracktor-no-FPN++ applies vanilla Faster R-CNN.}
\label{tab:mot_frcnn_data_association}

\end{table}

\subsection{Tracktor thresholds}
To demonstrate the robustness of our tracker with respect to the classification score and IoU thresholds, we refrained from any sequence or detection-specific fine-tuning.
In particular, we performed our experiments on all benchmarks with $\sigma_{active}=0.5$, $\lambda_{active}=0.6$ and $\lambda_{new}=0.3$, which were chosen to be optimal for the MOT17 training dataset.
In general, a higher $\lambda_{active}$ than $\lambda_{new}$ introduces stability into the tracker, as less active tracks are killed by the NMS and less new tracks are initialized.
A comparatively higher $\lambda_{active}$ relaxes potential object-object occlusions and implies a certain confidence in the regression performance.



\subsection{Tracktor video frame rate robustness}
A successful Tracktor bounding box regression depends on sufficiently high video frame rates or, in other words, small frame-by-frame object displacements.
A possible approach to address this issue is the extension with a powerful motion model.
A rudimentary motion model, the camera motion compensation (CMC), is presented in Section 2.3 and evaluated in the ablation study in Table 1. 
However, MOT16 and MOT17 mostly consist of sequences with benevolent video frame rates and slow moving objects (pedestrians).

We therefore complement our analysis of Tracktor in challenging tracking scenarios from Section 4.1 with an evaluation of its video frame rate robustness.
To this end, we evaluate Tracktor and Tracktor++ on all MOT17 training sequences with originally 30 frames per second (FPS) and reduce their frame rates by removing frames from the data and ground truth.
In Figure~\ref{fig:mot17_low_fps_tracktor}, both versions exhibit a fairly robust object tracking (MOTA) and identity preservation (IDF1) for rates as low as 5 FPS.
As expected, the performance for very small rates suffers particularly with respect to identity preservation.


\begin{figure}[h]
    \centering
    \includegraphics[width=\columnwidth]{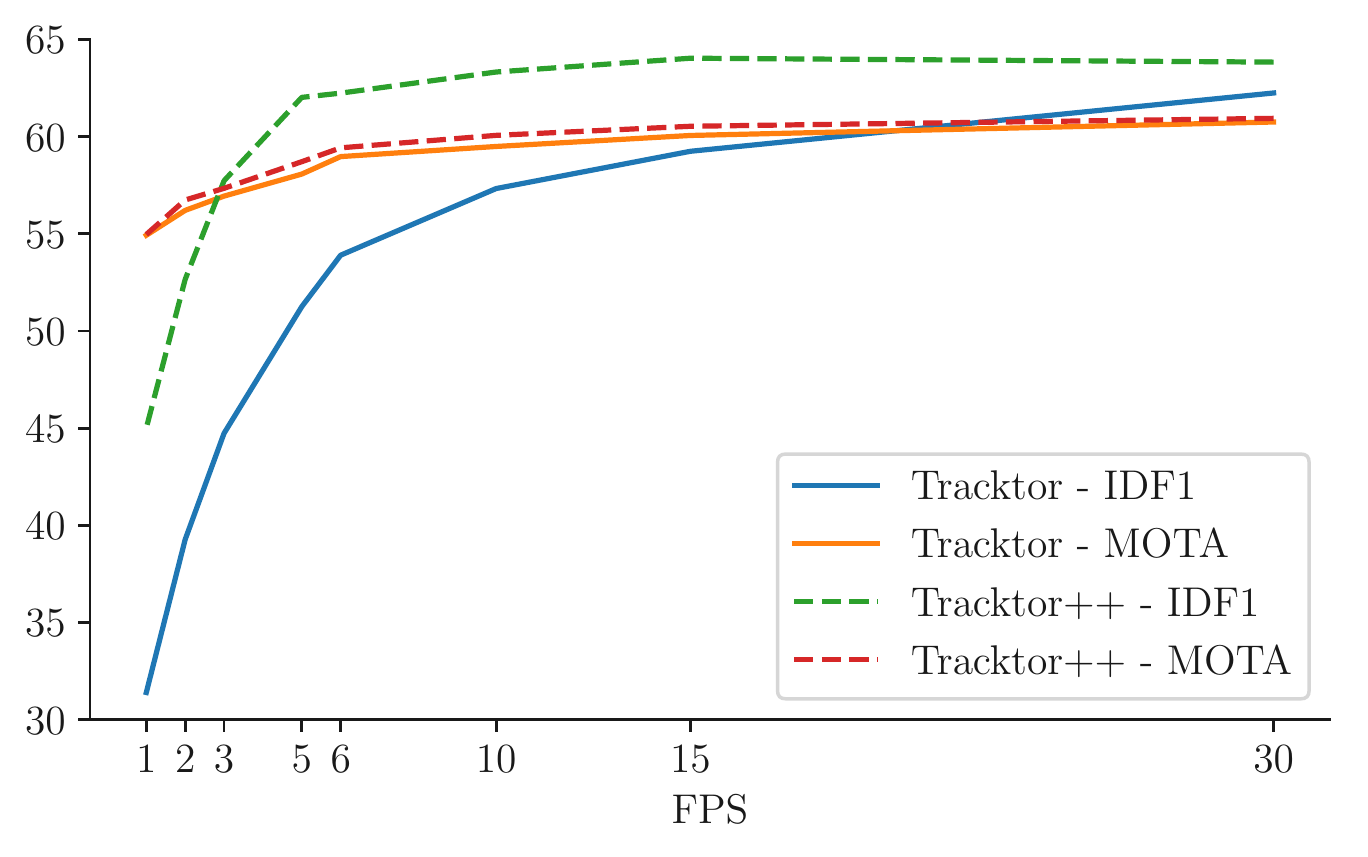}
    \vspace*{-7mm}
    \caption{Tracking performance of Tracktor and Tracktor++ on low frame rate versions of the MOT17-\{02, 04, 09, 10, 11\}-FRCNN sequences.}
    \label{fig:mot17_low_fps_tracktor}
\end{figure}

\section{Oracle trackers}

In our main work, we conclude the analysis in Section~\ref{sec:analysis} with a comparison of multiple oracle trackers that highlight the potential of future research directions.
For each oracle, one or multiple aspects of our vanilla Tracktor are substituted with ground truth information, thereby simulating perfect behavior.
For further understanding, we provide more details on the oracles for each of the distinct tracking aspects:

\begin{itemize}
    \item \textbf{Oracle-Kill:} This oracle kills tracks only if they have an IoU less than 0.5 with the corresponding ground truth bounding box.
    The matching between predicted and ground truth tracks is performed with the Hungarian~\cite{Kuhn55thehungarian} algorithm. 
    In the case of an object-object occlusion (IoU $> 0.8$), the ground truth matching is applied to decide which of the objects is occluded by the other and therefore should be killed.
    

    \item \textbf{Oracle-REG:} We simulate a perfect regression by matching tracks with an IoU threshold of 0.5 to the ground truth at frame $t-1$ .
    The regression oracle then sets track bounding boxes to the corresponding ground truth coordinates at frame $t$.


    \item \textbf{Oracle-MM:} A perfect motion model works analogous to Oracle-REG but we only move the previous bounding box center to the center of the ground truth bounding box at frame $t$.
    However, the bounding box height and width are still determined by the regression.
    

    \item \textbf{Oracle-reID:} Again, we use the Hungarian algorithm to match the new set of detections to the ground truth data. 
    Ground truth identity matches between inactive tracks and new detections yield a perfect re-identification.
\end{itemize}

\clearpage


\begin{algorithm}[h!]
    \KwData{Video sequence as ordered list $I = \{i_0, i_1, \cdots, i_{T-1}\}$  of images $i_t$.}
    \vspace*{24px}
    \KwResult{Set of object trajectories $\mathcal{T} = \{T_1, T_1, \cdots, T_k\}$ with $T_k=\{{{\bf b}^k_{t_1}},{{\bf b}^k_{t_2}}, \cdots , {{\bf b}^k_{t_N}} \, | \, 0 \leq t_1, \cdots, t_N \leq T-1\}$ as a list of ordered object bounding boxes ${\bf b}^k_t=(x,y,w,h)$.}

    \SetKwFunction{nms}{NMS}
    \SetKwFunction{zip}{zip}
    \SetKwFunction{iou}{IoU}

    \SetKwData{Ta}{$\mathcal{T}_\mathrm{active}$}
    \SetKwData{T}{$\mathcal{T}$}
    \SetKwData{ssr}{$\sigma_\mathrm{active}$}
    \SetKwData{ssd}{$\sigma_\mathrm{new}$}
    \SetKwData{la}{$\lambda_\mathrm{active}$}
    \SetKwData{ln}{$\lambda_\mathrm{new}$}
    \SetKwData{Sa}{$S_\mathrm{active}$}
    \SetKwData{D}{$\mathcal{D}_t$}
    \SetKwData{Sd}{$S_\mathrm{det}$}

    $\T, \Ta   \leftarrow  \emptyset$\;
    
    \For{$i_t \in I$}{
        $B, S   \leftarrow  \emptyset$\;
        \For{$T_k \in \Ta$}{
            $\mathrm{\mathbf{b}}_{t-1}^k \leftarrow T_k\lbrack-1\rbrack$\;
            $\mathbf{b}_t^k, s_t^k \leftarrow \mathrm{detector.reg\_and\_class}(\mathbf{b}_{t-1}^k)$\;
            
            \eIf{$\mathrm{s}_t^k < \ssr$}
                {
                $\Ta \leftarrow \Ta - \{T_k\}$\;
                $\T \leftarrow \T + \{T_k\}$\;}
                {
                $B \leftarrow B + \{\mathbf{b}_t^k\}$\;
                $S \leftarrow S + \{\mathrm{s}_t^k\}$\;
                }}

        $B \leftarrow \nms{B, S, \la}$\;

        \For{$k, T_k \in \Ta$}
        {
            \If{$k \notin B$}
            {
                $\Ta \leftarrow \Ta - \{T_k\}$\;
                $\T \leftarrow \T + \{T_k\}$\;
            }
        }

        \For{$T_k, \mathbf{b}_t^k  \in \zip{\Ta, B}$}
        {
            $T_k \leftarrow T_k + \{\mathbf{b}_t^k\}$\;
        }

        $\mathcal{D}_t \leftarrow \mathrm{detector.detections}(i_t)$\;
        \For{$d_t \in \mathcal{D}_t$}{
            \For{$\mathbf{b}_t^k \in B$}{
                \If{$\iou{$d_t, \mathbf{b}_t^k$} > \ln$}{
                    $\mathcal{D}_t \leftarrow \mathcal{D}_t - \{d_t\}$\;}}}

        \For{$d_t \in \mathcal{D}_t$}{
            $T_k  \leftarrow  \emptyset$\;
            $T_k \leftarrow T_k + \{d_t\}$\;
            $\Ta \leftarrow \Ta + \{T_k\}$\;
        }

        }
    $\T \leftarrow \T + \Ta$\; 
 
\caption{Tracktor algorithm (private detections)}
\label{alg:method_tracktor_private}
\end{algorithm}
\newpage

\begin{algorithm}[h!]
    \KwData{Video sequence as ordered list $I = \{i_0, i_1, \cdots, i_{T-1}\}$  of images $i_t$ and public detections as ordered list $\mathcal{D} = \{\mathcal{D}_0,\mathcal{D}_1, \cdots, \mathcal{D}_{T-1}\}$ of detections $\mathcal{D}_t$.}
    \KwResult{Set of object trajectories $\mathcal{T} = \{T_1, T_2 \cdots, T_k\}$ with $T_k=\{{{\bf b}^k_{t_1}},{{\bf b}^k_{t_2}}, \cdots , {{\bf b}^k_{t_N}} \, | \, 0 \leq t_1, \cdots, t_N \leq T-1\}$ as a list of ordered object bounding boxes ${\bf b}^k_t=(x,y,w,h)$.}

    \SetKwFunction{nms}{NMS}
    \SetKwFunction{zip}{zip}
    \SetKwFunction{iou}{IoU}

    \SetKwData{Ta}{$\mathcal{T}_\mathrm{active}$}
    \SetKwData{T}{$\mathcal{T}$}
    \SetKwData{ssr}{$\sigma_\mathrm{active}$}
    \SetKwData{ssd}{$\sigma_\mathrm{new}$}
    \SetKwData{la}{$\lambda_\mathrm{active}$}
    \SetKwData{ln}{$\lambda_\mathrm{new}$}
    \SetKwData{Sa}{$S_\mathrm{active}$}
    \SetKwData{D}{$\mathcal{D}$}
    \SetKwData{Dt}{$\mathcal{D}_t$}
    \SetKwData{Sd}{$S_\mathrm{det}$}

    $\T, \Ta   \leftarrow  \emptyset$\;
    
    \For{$i_t, \Dt \in \zip{I, \D}$}{
        $B, S   \leftarrow  \emptyset$\;
        \For{$T_k \in \Ta$}{
            $\mathrm{\mathbf{b}}_{t-1}^k \leftarrow T_k\lbrack-1\rbrack$\;
            $\mathbf{b}_t^k, s_t^k \leftarrow \mathrm{detector.reg\_and\_class}(\mathbf{b}_{t-1}^k)$\;
            
            \eIf{$\mathrm{s}_t^k < \ssr$}
                {
                $\Ta \leftarrow \Ta - \{T_k\}$\;
                $\T \leftarrow \T + \{T_k\}$\;}
                {
                $B \leftarrow B + \{\mathbf{b}_t^k\}$\;
                $S \leftarrow S + \{\mathrm{s}_t^k\}$\;
                }}

        $B \leftarrow \nms{B, S, \la}$\;

        \For{$k, T_k \in \Ta$}
        {
            \If{$k \notin B$}
            {
                $\Ta \leftarrow \Ta - \{T_k\}$\;
                $\T \leftarrow \T + \{T_k\}$\;
            }
        }

        \For{$T_k, \mathbf{b}_t^k  \in \zip{\Ta, B}$}
        {
            $T_k \leftarrow T_k + \{\mathbf{b}_t^k\}$\;
        }

        $S \leftarrow  \emptyset$\;
        \For{$\mathbf{d}_t \in \Dt$}{
            $\mathbf{d}_t, s_t \leftarrow \mathrm{detector.reg\_and\_class}(\mathbf{d}_t)$\;
            \eIf{$s_t < \ssr$}
            {
                $\Dt \leftarrow \Dt - \{\mathbf{d}_t\}$\;
            }
            {
                $S \leftarrow S + \{s_t\}$\;
            }
        }
        $\Dt \leftarrow \nms{\Dt, S, \ln}$\;
        \For{$\mathbf{d}_t \in \Dt$}{
            \For{$\mathbf{b}_t^k \in B$}{
                \If{$\iou{$d_t, \mathbf{b}_t^k$} > \ln$}{
                    $\mathcal{D}_t \leftarrow \mathcal{D}_t - \{d_t\}$\;}}
        }

        \For{$d_t \in \mathcal{D}_t$}{
            $T_k  \leftarrow  \emptyset$\;
            $T_k \leftarrow T_k + \{d_t\}$\;
            $\Ta \leftarrow \Ta + \{T_k\}$\;
        }

        }
    $\T \leftarrow \T + \Ta$\; 
 
\caption{Tracktor algorithm (public detections)}
\label{alg:method_tracktor_public}
\end{algorithm}

\begin{table*}[p]
\center
\tabcolsep=0.11cm
    \resizebox{1.5\columnwidth}{!}{
    \begin{tabular}{l l c c c c c c c}
     \toprule
     Sequence & Detection & MOTA $\uparrow$ & IDF1 $\uparrow$ & MT $\uparrow$ & ML $\downarrow$ & FP $\downarrow$ & FN $\downarrow$ & ID Sw. $\downarrow$ \\ [0.5ex] 
     \midrule
     \multicolumn{8}{c}{MOT17~\cite{milanarxiv2016}} \\
     \midrule
     MOT17-01 & DPM~\cite{dpmpami2009} & 35.9 & 37.1 & 20.8 & 50.0 & 131 & 3962 & 39 \\
     MOT17-03 & DPM & 65.2 & 57.0 & 35.1 & 12.8 & 1338 & 34840 & 222 \\
     MOT17-06 & DPM & 52.7 & 55.7 & 18.5 & 40.1 & 184 & 5310 & 80 \\
     MOT17-07 & DPM & 40.5 & 42.5 & 10.0 & 40.0 & 363 & 9603 & 90 \\
     MOT17-08 & DPM & 27.0 & 30.7 & 9.2 & 50.0 & 213 & 15130 & 83 \\
     MOT17-12 & DPM & 45.6 & 55.2 & 16.5 & 48.4 & 88 & 4596 & 29 \\
     MOT17-14 & DPM & 26.9 & 37.1 & 6.7 & 53.0 & 591 & 12834 & 92 \\
     \midrule
     MOT17-01 & FRCNN~\cite{rennips2015} & 34.9 & 34.8 & 20.8 & 41.7 & 406 & 3753 & 39 \\
     MOT17-03 & FRCNN & 66.4 & 59.7 & 37.2 & 13.5 & 1014 & 33961 & 189 \\
     MOT17-06 & FRCNN & 56.7 & 59.0 & 23.0 & 27.5 & 359 & 4647 & 96 \\
     MOT17-07 & FRCNN & 39.4 & 43.1 & 11.7 & 40.0 & 555 & 9588 & 93 \\
     MOT17-08 & FRCNN & 27.1 & 31.7 & 11.8 & 50.0 & 197 & 15119 & 74 \\
     MOT17-12 & FRCNN & 43.4 & 53.9 & 15.4 & 51.6 & 185 & 4697 & 25 \\
     MOT17-14 & FRCNN & 27.1 & 38.1 & 7.3 & 48.2 & 1202 & 12139 & 132 \\
     \midrule
     MOT17-01 & SDP~\cite{sdpYangcvpr2016} & 37.5 & 36.8 & 25.0 & 41.7 & 283 & 3706 & 42 \\
     MOT17-03 & SDP & 69.6 & 60.1 & 39.9 & 10.8 & 2469 & 29065 & 248 \\
     MOT17-06 & SDP & 56.8 & 59.2 & 26.1 & 28.8 & 354 & 4638 & 93 \\
     MOT17-07 & SDP & 41.2 & 42.6 & 11.7 & 33.3 & 596 & 9231 & 111 \\
     MOT17-08 & SDP & 28.7 & 32.1 & 13.2 & 47.4 & 253 & 14715 & 103 \\
     MOT17-12 & SDP & 45.3 & 56.9 & 18.7 & 48.4 & 212 & 4492 & 34 \\
     MOT17-14 & SDP & 27.6 & 38.5 & 7.3 & 48.2 & 1208 & 12021 & 158 \\
     \midrule
     \multicolumn{2}{c}{All} & 53.5 & 52.3 & 19.5 & 36.6 & 12201 & 248047 & 2072\\
     \midrule
     \multicolumn{8}{c}{MOT16~\cite{milanarxiv2016}} \\
     \midrule
     MOT16-03 & DPM & 65.8 & 57.9 & 35.1 & 12.2 & 1397 & 34101 & 226 \\
     MOT16-06 & DPM & 53.9 & 57.9 & 20.4 & 39.4 & 243 & 5000 & 80 \\
     MOT16-07 & DPM & 43.0 & 43.6 & 13.0 & 33.3 & 405 & 8808 & 97 \\
     MOT16-08 & DPM & 34.3 & 36.8 & 12.7 & 38.1 & 314 & 10577 & 101 \\
     MOT16-12 & DPM & 48.0 & 57.0 & 18.6 & 44.2 & 108 & 4172 & 30 \\
     MOT16-14 & DPM & 27.4 & 37.6 & 6.7 & 51.2 & 659 & 12645 & 108 \\
     \midrule
     \multicolumn{2}{c}{All} & 54.4 & 52.5 & 19.0 & 36.9 & 3280 & 79149 & 682 \\
     \midrule
     \multicolumn{8}{c}{2D MOT 2015~\cite{lealarxiv2015}} \\
     \midrule
     TUD-Crossing & ACF~\cite{acfpami2014} & 78.3 & 58.3 & 53.8 & 0.0 & 14 & 207 & 18 \\
     PETS09-S2L2 & ACF & 44.5 & 28.4 & 4.8 & 2.4 & 644 & 4420 & 289 \\
     ETH-Jelmoli & ACF & 57.8 & 67.4 & 35.6 & 24.4 & 317 & 732 & 21 \\
     ETH-Linthescher & ACF & 49.3 & 55.5 & 15.7 & 50.8 & 178 & 4303 & 48 \\
     ETH-Crossing & ACF & 43.0 & 54.2 & 11.5 & 38.5 & 22 & 538 & 12 \\
     AVG-TownCentre & ACF & 39.0 & 38.5 & 17.3 & 19.0 & 620 & 3075 & 665 \\
     ADL-Rundle & ACF-1 & 33.7 & 49.3 & 28.1 & 9.4 & 2497 & 3615 & 56 \\
     ADL-Rundle & ACF-3 & 45.6 & 46.0 & 15.9 & 13.6 & 750 & 4713 & 68 \\
     KITTI-16 & ACF & 48.1 & 50.8 & 17.6 & 5.9 & 174 & 672 & 37 \\
     KITTI-19 & ACF & 49.4 & 59.5 & 14.5 & 14.5 & 553 & 2082 & 71 \\
     Venice-1 & ACF & 35.1 & 42.6 & 23.5 & 29.4 & 708 & 2220 & 33 \\
     \midrule
     \multicolumn{2}{c}{All} & 44.1 & 46.7 & 18.0 & 26.2 & 6477 & 26577 & 1318 \\
     \bottomrule
    \end{tabular}}

\caption{A detailed summary of the tracking results of our Tracktor++ tracker on all three MOTChallenge benchmarks. The results are separated into individual sequences and sets of public detections.}
\label{tab:tracktorplusplus_verbose}

\end{table*}
\fi

\clearpage

{\small
\bibliographystyle{ieee_fullname}
\bibliography{pubs,cvpr2019,refs-alt}
}

\end{document}